\documentclass{article}
\usepackage[preprint]{neurips_2025}
\usepackage[pagebackref=true,breaklinks=true,colorlinks,urlcolor=black,bookmarks=false]{hyperref}
\hypersetup{citecolor=[RGB]{119,185,0}}

\usepackage{amsmath}  
\usepackage{amssymb}  
\usepackage{algorithmicx,algorithm}
\usepackage{amsmath}
\usepackage{xspace}
\usepackage[utf8]{inputenc} % allow utf-8 input
\usepackage[T1]{fontenc}    % use 8-bit T1 fonts
\usepackage{hyperref}       % hyperlinks
\usepackage{url}            % simple URL typesetting
\usepackage{booktabs}       % professional-quality tables
\usepackage{amsfonts}       % blackboard math symbols
\usepackage{nicefrac}       % compact symbols for 1/2, etc.
\usepackage{footmisc}
\usepackage[nopatch=footnote]{microtype}
\usepackage{xcolor}         % colors
\usepackage{multirow}
\usepackage{array}       
\usepackage{tabularx}    
\usepackage{algpseudocode}                  % algorithm formatting
\usepackage{setspace}                       % pleasant display of algorithm.
\usepackage{lipsum} % 
\usepackage{subcaption} % subtable
\usepackage{enumitem}
\usepackage{caption} % captionof for minipage
\usepackage{colortbl} % rowcolor
\usepackage{bbm}
\usepackage{algorithm}
\usepackage{graphicx}
\usepackage{xcolor}
\usepackage{relsize}
\usepackage{etoolbox}
\usepackage{wrapfig} % for wraptable environment

\newcommand{\sname}{REG\xspace}
\newcommand{\pz}

\definecolor{lightpurple}{RGB}{223,215,229}
\definecolor{lightbule}{RGB}{213,212,230}
\definecolor{lightpink}{RGB}{240,210,217}
\definecolor{lightgreen}{RGB}{208,232,234}
\definecolor{tabhighlight}{HTML}{e5e5e5}

\definecolor{zhenyuanorange}{RGB}{255,140,0}

\newcommand{\maketitlesupplementary}{
  \begin{center}
    \Large{Representation Entanglement for Generation:}\\
    \Large{Training Diffusion Transformers Is Much Easier Than You Think}\\
    \Large{Supplementary Materials} 
  \end{center}
}

\title{
  \makebox[\textwidth][c]{Representation Entanglement for Generation:} \\
  \makebox[\textwidth][c]{Training Diffusion Transformers Is Much Easier Than You Think}
}

\author{
  Ge Wu$^{1,2}$\ 
  \quad Shen Zhang$^{3}$\ 
  \quad Ruijing Shi$^{1}$\ 
  \quad Shanghua Gao$^{4}$\ 
  \quad Zhenyuan Chen$^{1,2}$\  
  \quad Lei Wang$^{1,2}$\ \\ \textbf{Zhaowei Chen$^{3}$\ \quad Hongcheng Gao$^{5}$\ \quad Yao Tang$^{3}$\ \quad Jian Yang$^{1}$ }  \textbf{ Ming-Ming Cheng$^{1,2}$\ \quad Xiang Li$^{1,2}$\footnotemark[2]\ }
    \\[0.04cm]
    $^{1}$NKIARI, Shenzhen Futian,
    $^{2}$VCIP, CS, Nankai University,
    $^{3}$JIIOV Technology, \\
    $^{4}$Harvard University,
    $^{5}$University of Chinese Academy of Sciences
}

\begin{document}

\maketitle

\renewcommand{\thefootnote}{\fnsymbol{footnote}}

\footnotetext[2]{corresponding author. xiang.li.implus@nankai.edu.cn}

\vspace{-12mm}

\begin{figure}[ht]
    \centering
    \includegraphics[width=0.9\linewidth]{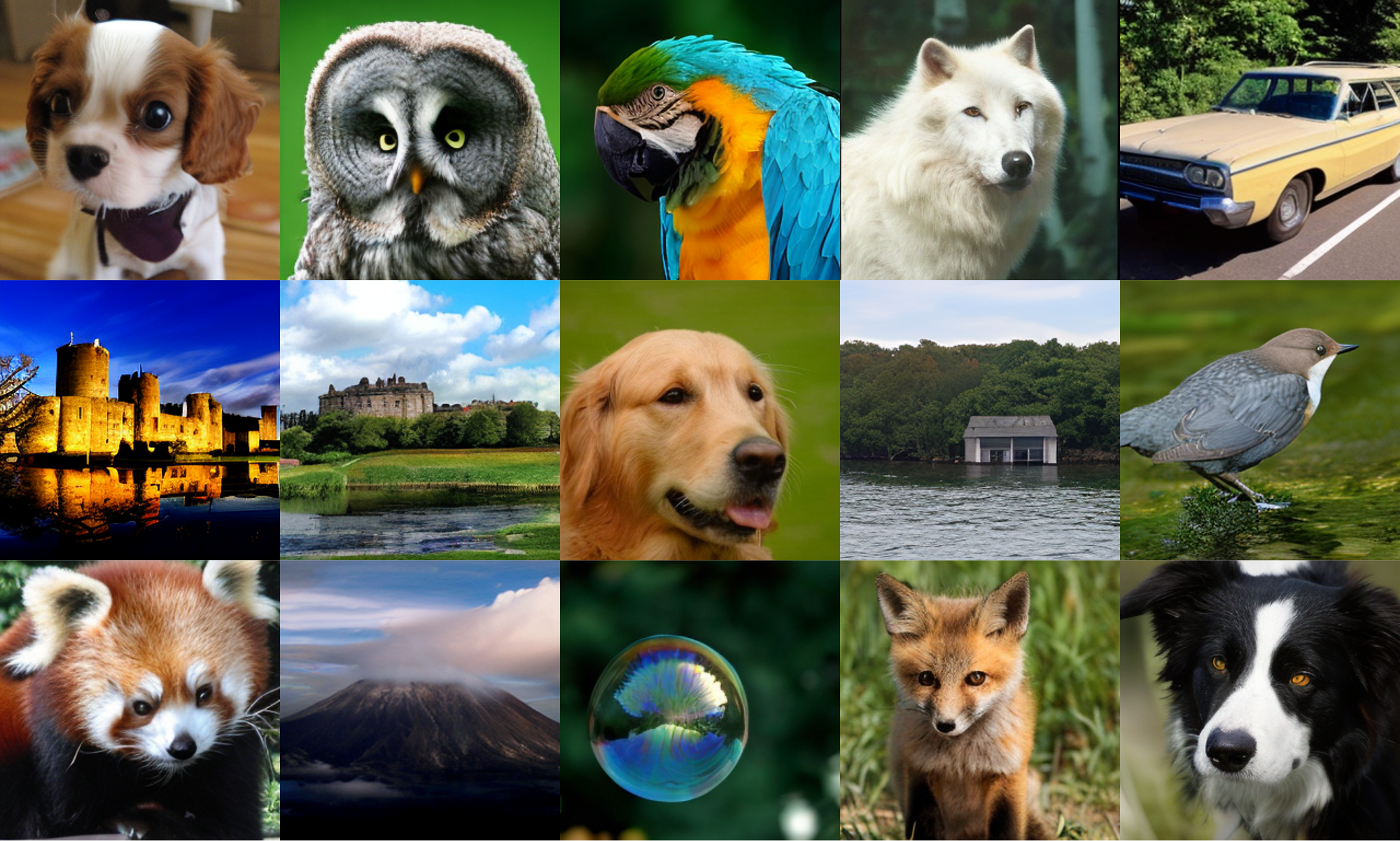}
    %\captionsetup{font={footnotesize}}
    \vspace{-5pt}
    \caption{\textbf{Representation Entanglement for Generation demonstrates excellent image quality.} }
    \label{fig:abstract-fig}
    \vspace{-1em}
\end{figure}

\begin{abstract}
REPA and its variants effectively mitigate training challenges in diffusion models by incorporating external visual representations from pretrained models, through alignment between the noisy hidden projections of denoising networks and foundational clean image representations. We argue that the external alignment, which is absent during the entire denoising inference process, falls short of fully harnessing the potential of discriminative representations. In this work, we propose a straightforward method called \textit{\textbf{R}epresentation \textbf{E}ntanglement for \textbf{G}eneration} (\textbf{REG}), which entangles low-level image latents with a single high-level class token from pretrained foundation models for denoising. 
REG acquires the capability to produce coherent image-class pairs directly from pure noise, substantially improving both generation quality and training efficiency.
This is accomplished with negligible additional inference overhead, requiring only one single additional token for denoising (<0.5\% increase in FLOPs and latency).
The inference process concurrently reconstructs both image latents and their corresponding global semantics, where the acquired semantic knowledge actively guides and enhances the image generation process.
On ImageNet 256$\times$256, SiT-XL/2 + REG demonstrates remarkable convergence acceleration, achieving $\textbf{63}\times$ and $\textbf{23}\times$ faster training than SiT-XL/2 and SiT-XL/2 + REPA, respectively. 
More impressively, SiT-L/2 + REG trained for merely 400K iterations outperforms SiT-XL/2 + REPA trained for 4M iterations ($\textbf{10}\times$ longer). Code is available at: \url{https://github.com/Martinser/REG}. 
\end{abstract}

\begin{figure*}[t]
    \centering
    \includegraphics[width=1.0\linewidth]{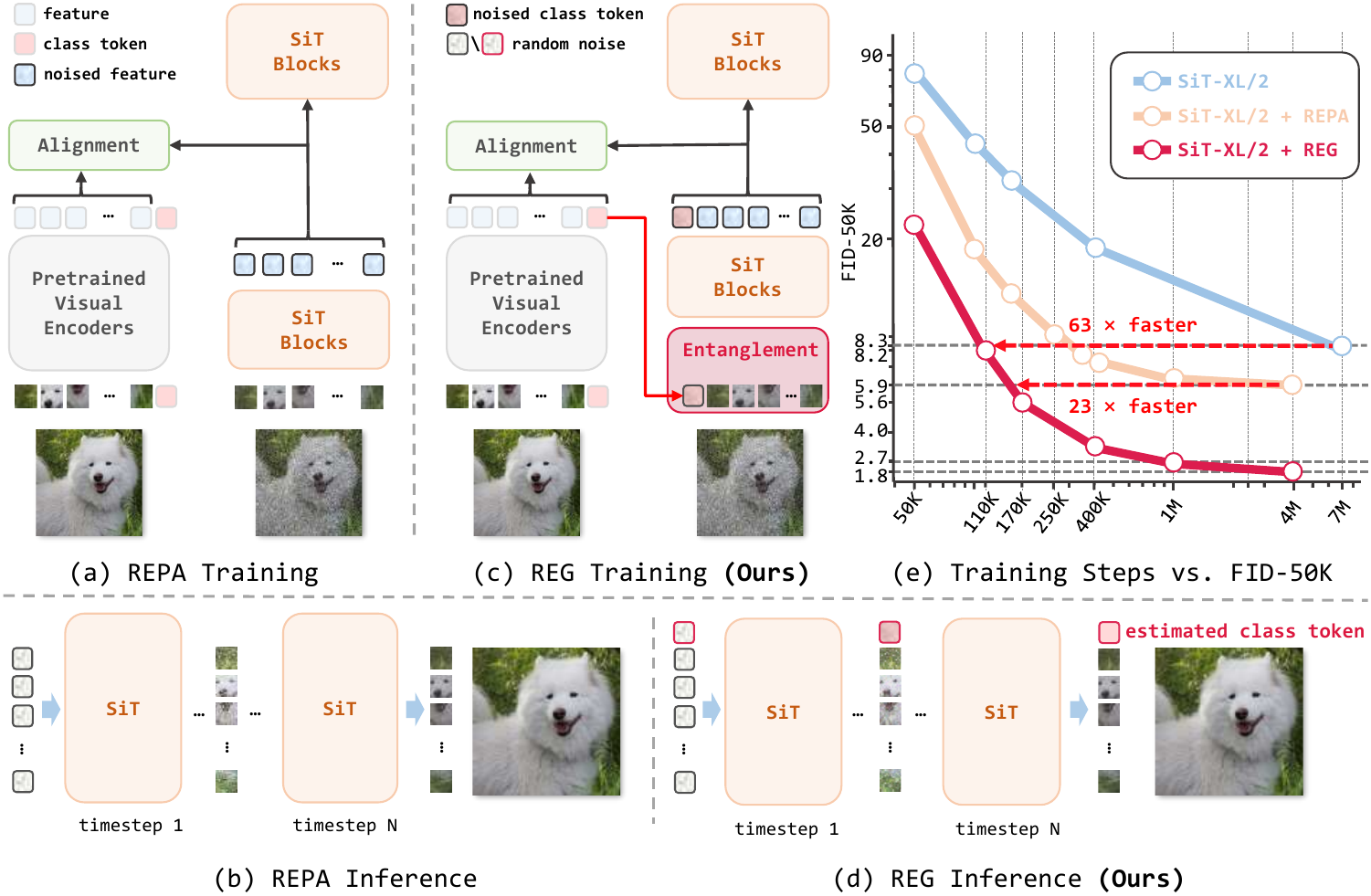}
    \caption
    {
        \textbf{Comparison between REPA and our Representation Entanglement for Generation (REG).}
        (a) During training, REPA~\cite{repa} indirectly aligns the intermediate denoising features of SiT~\cite{sit} with pretrained foundation model representations during training.
        (b) The external alignment of REPA is absent during actual denoising inference, limiting the effectiveness of discriminative information.
        (c) REG entangles low-level image latents with a pretrained foundation model's class token in training, providing discriminative semantic guidance to SiT. 
        (d) REG's inference process jointly reconstructs both image latents and their associated global semantics from random noise initialization. The incorporated semantic knowledge continues to guide generation, actively enhancing image quality during inference.
        (e) On ImageNet 256$\times$256, SiT-XL/2 + REG achieves substantial acceleration in convergence, training \textbf{63}$\times$ and \textbf{23}$\times$ faster than SiT-XL/2 and SiT-XL/2 + REPA, respectively.
    }
    \label{fig:angle_resolver}
    \vspace{-10pt}
\end{figure*}

\section{Introduction}

Generative models have undergone significant evolution~\cite{vqgan,VAR,ho2020denoising,flowgen,DALLE}, demonstrating remarkable success across diverse applications~\cite{hu2024token,hassan2024gem,liang2018jtav,nijkamp2022codegen,zhang2025ledit,hu2025anchor,liu2025cradle}. 
Recent progress in high-fidelity image synthesis has been driven by several key innovations: Latent Diffusion Models (LDM)~\cite{ldm} introduced a stable two-phase training framework, while Diffusion Transformers (DiT)~\cite{dit} enhanced scalability through transformer-based architectures. Building upon these developments, Scalable Interpolant Transformers (SiT)~\cite{sit} further unified the approach through continuous-time stochastic interpolants for diffusion training.
Despite these advances, achieving high-fidelity synthesis remains a substantial resource for model convergence. While recent techniques such as masked training~\cite{mdt,mdtv2} and multi-scale optimization~\cite{dimr} partially alleviate computational costs and accelerate model convergence. However, fundamental optimization challenges persist when relying solely on architecture changes.

Recent studies demonstrate that enhanced generative models can acquire more discriminative representations, positioning them as capable representation learners~\cite{prof,repa,repa-e}. However, as quantified by CKNNA metrics~\cite{cknna}, these features still underperform compared to those from pretrained vision models~\cite{dinov2, mocov3, clip}. 
This performance gap has motivated approaches leveraging pretrained visual encoder features to accelerate generative model training convergence. For example, REPA~\cite{repa} employs implicit feature-space alignment between diffusion models and foundation vision models (see Fig.~\ref{fig:angle_resolver}(a)), while REPA-E~\cite{repa-e} extends this alignment by enabling end-to-end VAE tuning, and quantitatively demonstrates that enhanced alignment (via increased CKNNA scores directly improves generation fidelity.)
However, the external alignment of REPA, which is absent during the entire denoising inference process, falls short of fully harnessing the potential of discriminative information (see Fig.~\ref{fig:angle_resolver}(b)). We suggest this structure likely impedes further advancements in discriminative semantic learning and overall generative capability.

To address these limitations, we propose a straightforward method called \textit{\textbf{R}epresentation \textbf{E}ntanglement for \textbf{G}eneration} (\textbf{REG}), an efficient framework that unleashes the potential of discriminative information through explicitly reflows discriminative information into the generation process (see Fig.~\ref{fig:angle_resolver}(c)). 
REG entangles low-level image latents with a single high-level class token from pretrained foundation models during training by applying synchronized noise injection to both of them with spatial concatenation. 
The denoising inference process concurrently reconstructs both image latents and their corresponding global semantics from random noise initialization, where the acquired semantic knowledge actively guides and enhances the image generation process (see Fig.~\ref{fig:angle_resolver}(d)).
REG achieves significant improvements in generation quality, training convergence speed, and discriminative semantic learning, all while introducing minimal computational cost through the addition of just one token (less than 0.5\% FLOPs and latency in Tab.~\ref{tab:flops}).
On class-conditional ImageNet benchmarks at 256$\times$256 resolution (see Fig.~\ref{fig:angle_resolver}(e)), SiT-XL/2 + REG achieves 63$\times$ and 23$\times$ faster training convergence compared to SiT-XL/2 and SiT-XL/2 + REPA, respectively. Notably, SiT-L/2 + REPA trained for 400K iterations surpasses the performance of SiT-XL/2 + REPA trained for 4M iterations (see Tab.~\ref{tab:wo_cfg}).

In summary, our specific contributions are as follows:\begin{itemize}[]
    \item[$\bullet$] We propose \textbf{REG}, an efficient framework that entangles low-level image latents with a single high-level class token from pretrained foundation models for denoising.
    \item[$\bullet$] REG significantly enhances generation quality, training convergence speed, and discriminative semantic learning while introducing negligible computational overhead.
    \item[$\bullet$] On ImageNet generation benchmarks, REG achieves 63$\times$ and 23$\times$ faster training convergence than SiT and REPA.
\end{itemize}

\section{Related work}

\textbf{Generative models for image generation.} Traditional approaches such as DDPM~\cite{ho2020denoising} and DDIM~\cite{ddim} perform iterative noise removal in pixel space, while LDM~\cite{ldm} operates in compressed latent spaces through pretrained autoencoders. Architecturally, early U-Net-based diffusion models~\cite{ho2020denoising,nichol2021improved,ldm} rely on iterative denoising, whereas modern transformer-based frameworks like DiT~\cite{dit} and SiT~\cite{sit} leverage self-attention mechanisms for superior spatial pattern modeling. Despite these advances, existing methods typically require extensive training iterations to achieve convergence.
Current acceleration techniques often necessitate significant architectural modifications, such as masked training paradigms~\cite{mdt,mdtv2} or multi-scale optimization strategies~\cite{piip,dimr}. In contrast, we propose REG, which achieves dual improvements in generation quality and training efficiency while introducing minimal inference overhead (requiring just one additional token during denoising). Crucially, REG accomplishes these gains while preserving the original model architecture, demonstrating that superior training dynamics can be achieved without structural compromises.

\textbf{Generative models as representation learners.} Extensive research has established that intermediate features in diffusion models inherently encode rich semantic representations~\cite{repa,repa-e}, with demonstrated discriminative capabilities across diverse vision tasks including semantic segmentation~\cite{Segmentation1,Segmentation2,Segmentation3}, depth estimation~\cite{depth1}, and controllable image editing~\cite{edit1,edit2,edit3}. Recent advancements have further developed knowledge transfer paradigms from diffusion models to efficient networks through techniques like RepFusion's dynamic timestep optimization~\cite{repdiff} and DreamTeacher's cross-model feature distillation~\cite{dreamteacher}.
Notably, DDAE~\cite{prof} confirms that improved diffusion models yield higher-quality representations, establishing a direct correlation between generation capability and representation learning performance. Building upon these insights, we propose to systematically integrate discriminative representations into the generative forward process, enabling persistent discriminative guidance throughout denoising inference.

\textbf{Generative models with external representations.} Prior research~\cite{extra1,extra2,vavae} has explored augmenting diffusion models through auxiliary components. For example, RCG~\cite{rcg} employs a secondary diffusion model to generate the class token for adaLN-condition~\cite{adaln} in unconditional generation. In contrast, our approach eliminates the need for additional models by leveraging a single class token as part of the input to provide discriminative guidance, simultaneously enhancing both discriminative semantic learning and conditional generation performance.
Recent advancements have incorporated visual representations from foundation models to accelerate diffusion training. REPA~\cite{repa} improves semantic representation quality through feature alignment between early diffusion layers and pretrained vision features, while REPA-E~\cite{repa-e} extends this framework by enabling end-to-end VAE tuning. 
% However, these methods rely on external alignment mechanisms that remain inactive during denoising inference, fundamentally limiting their capacity to fully exploit discriminative representations.
However, these methods rely on external alignment mechanisms that do not take the discriminative representations as the input and denoising, which are unable to produce discriminative representations during inference to guide the generation process.
Our proposed REG framework structurally integrates spatial visual representations with semantic class embeddings derived from foundation models. This architectural design enables the denoising phase to concurrently refine localized pattern restoration and holistic conceptual representation, thereby establishing context-aware semantic steering that persists throughout the entire generative process.

\section{Method}

We propose \textbf{REG}, an efficient framework that provides discriminative guidance by entangling image latents with foundation model class token (Fig.~\ref{fig:angle_resolver}(c, d)). Section~\ref{sec:3.1} covers preliminaries, followed by REG's detailed description in Section~\ref{sec:3.2}.

\subsection{Preliminaries}
\label{sec:3.1}
Our work is based on Scalable Interpolant Transformers~(SiT)~\cite{sit}, which provide a unified perspective to understand flow and diffusion models. We first introduce the relevant preliminaries.
Flow and diffusion models both leverage stochastic processes to gradually transform Gaussian noise $\epsilon \sim \mathcal{N}(0, I)$ into data samples $\mathbf{x}_*$. This process can be unified as
\begin{equation}
    \mathbf{x}_t = \alpha_t \mathbf{x}_* + \sigma_t \epsilon,
\end{equation}
where $\alpha_t$ is a decreasing and $\sigma_t$ an increasing function of time $t$. Flow-based models typically interpolate between noise and data over a finite interval, while diffusion models define a forward stochastic differential equation (SDE) that converges to a Gaussian distribution as $t \to \infty$.

Sampling from these models can be achieved via either a reverse-time SDE or a probability flow ordinary differential equation (ODE), both of which yield the same marginal distributions for $\mathbf{x}_t$. The probability flow ODE is:
\begin{equation}
    \dot{\mathbf{x}}_t = \mathbf{v}(\mathbf{x}_t, t),
\end{equation}
where the velocity field $\mathbf{v}(\mathbf{x}, t)$ can be formulated  by the conditional expectation:
\begin{equation} \label{velocity}
    \mathbf{v}(\mathbf{x}, t) = \mathbb{E}[\dot{\mathbf{x}}_t \mid \mathbf{x}_t = \mathbf{x}]
    = \dot{\alpha}_t \mathbb{E}[\mathbf{x}_* \mid \mathbf{x}_t = \mathbf{x}]
    + \dot{\sigma}_t \mathbb{E}[\epsilon \mid \mathbf{x}_t = \mathbf{x}].
\end{equation}
To synthesize data, we can integrate Eqn.~\eqref{velocity} in reverse time, initializing from random Gaussian noise $\epsilon \sim \mathcal{N}(0, \mathbf{I})$. This process yields samples from $p_0(\mathbf{x})$, serving as an approximation to the true data distribution $p(\mathbf{x})$.
This velocity can be estimated by a model $\mathbf{v}_\theta(\mathbf{x}_t, t)$, which is trained to minimize the following loss function:
% \begin{equation}
% \mathcal{L}_{\mathbf{v}}(\theta) = \int_{0}^{T} \mathbb{E} \left[ \left\| \mathbf{v}_{\theta}(\mathbf{x}_t, t) - \dot{\alpha}_t \mathbf{x}_* - \dot{\sigma}_t \boldsymbol{\epsilon} \right\|^2 \right] \mathrm{d}t.
% \end{equation}
\begin{equation}
\mathbb{E}_{\mathbf{x}_*,\boldsymbol{\epsilon},{t}}  [\left\| \mathbf{v}_{\theta}(\mathbf{x}_t, t) - \dot{\alpha}_t \mathbf{x}_* - \dot{\sigma}_t \boldsymbol{\epsilon} \right\|^2] .
\end{equation}

The reverse-time SDE can describe the probability distribution $p_t(\mathbf{x})$ of $\mathbf{x}_t$ at time $t$, which can be expressed as:
\begin{equation}
    d\mathbf{x}_t = \mathbf{v}(\mathbf{x}_t, t)dt - \frac{1}{2} w_t \mathbf{s}(\mathbf{x}_t, t)dt + \sqrt{w_t}d\overline{\mathbf{w}}_t,
\end{equation}

with $\mathbf{s}(\mathbf{x}, t)$ denoting the score  that can be computed via the conditional expectation:
\begin{equation}
    \mathbf{s}(\mathbf{x}_t, t) = -\sigma_t^{-1} \mathbb{E}[\epsilon \mid \mathbf{x}_t = \mathbf{x}].
    \label{eq:score_conditional}
\end{equation}
The score can be reformulated in terms of the velocity $\mathbf{v}(\mathbf{x}, t)$:
\begin{equation}
    \mathbf{s}(\mathbf{x}, t) = \sigma_t^{-1} \cdot \frac{\alpha_t \mathbf{v}(\mathbf{x}, t) - \dot{\alpha}_t \mathbf{x}}{\alpha_t \dot{\sigma}_t - \dot{\alpha}_t \sigma_t}.
    \label{eq:score_from_velocity}
\end{equation}
We can learn the velocity field $\mathbf{v}(\mathbf{x}, t)$ and use it to compute the score $\mathbf{s}(\mathbf{x}, t)$ when using an SDE for sampling.

\subsection{Representation entanglement for generation}
\label{sec:3.2}

\textbf{REG training process.} Given the clean input image $\mathbf{x}_*$, we obtain image latents $\mathbf{z}_{*}\in\mathbb{R}^{D_{z} \times C_z \times C_z}$ via VAE encoder~\cite{ldm} and image feature~$\mathbf{f}_{*}\in\mathbb{R}^{N \times D_{vf}}$ from vision foundation encoder (e.g. DINOv2~\cite{dinov2}), where $C_z \times C_z$ denotes the latent spatial resolution, and $D_{z}$ is the channel dimension. Besides, $N$ represents the number of visual tokens, and $D_{vf}$ is the embedding dimension of vision foundation encoder. 
% To address the problems of REPA, the external alignment is absent during the entire denoising inference process, falling short of fully harnessing the potential of discriminative information.
In REPA, the absence of the ability to autonomously generate discriminative representations to guide generation in inference may reduce the leverage of discriminative information effectively.
We introduce the class token $\mathbf{cls}_{*}\in \mathbb{R}^{1 \times D_{vf}}$ from the vision foundation model to entangle with image latents for providing the discriminative guidance. Here are the specific details:

We inject noise into both the class token and image latents as a paired input for the SiT forward process~\cite{sit}. Specifically, given two Gaussian noise samples $\epsilon_{z} \sim \mathcal{N}(0, \mathbf{I})$ and $\epsilon_{cls} \sim \mathcal{N}(0, \mathbf{I})$ with sizes $\mathbb{R}^{D_z \times C_z \times C_z}$ and $\mathbb{R}^{1 \times D_{vf}}$ respectively, we perform interpolation operations at continuous time $t~\in[0,1]$ as follows:
\begin{equation}
\mathbf{z}_t=\alpha_t \mathbf{z}_{*}+\sigma_t \epsilon_z; \quad \mathbf{cls}_t=\alpha_t \mathbf{cls}_{*}+\sigma_t \epsilon_{cls},
\end{equation}
This defines intermediate states~$\mathbf{z}_t$ (noised latents) and $\mathbf{cls}_t$ (noised class token) in the forward diffusion process, where $\alpha_t$ and $\sigma_t$ control the generation trajectory. 
Then, we patchify $\mathbf{z}_t$ into $\mathbf{z}_t' \in \mathbb{R}^{N \times {D}_{z}^{'}}$, ${D}_{z}^{'}$ is the embedding dimension.
Afterwards, the class token $\mathbf{cls}_t$ is projected into the same embedding space via a linear layer to obtain $\mathbf{cls}_t' \in \mathbb{R}^{1 \times {D}_{z}^{'}}$. Finally, we concatenate them to form $\mathbf{h}_t = [\mathbf{cls}_t'; \mathbf{z}_t'] \in \mathbb{R}^{(N+1) \times {D}_{z}^{'}}$, which serves as the input to the subsequent SiT blocks.
We perform alignment at specific transformer layers $n$, where $n=4$ for SiT-B/2 + REG and $n=8$ for all other variants, maintaining consistency with REPA.
Specifically, we align the projected hidden state feature \scalebox{0.9}{$h_\phi(H_t^{[n]}) \in \mathbb{R}^{(N+1) \times D_{vf}}$} with the reference representation $\mathbf{y}_{*} \in \mathbb{R}^{(N+1) \times D_{vf}}$ which is concatenated by $\mathbf{cls}_{*}$ and $\mathbf{f}_{*}$.
\scalebox{0.9}{$\mathbf{h}_t^{[n]} \in \mathbb{R}^{(N+1) \times {D}_{z}^{'}}$} denotes output of the $n$-th SiT block, $h_\phi$ is a trainable \textrm{MLP} projection, and $\operatorname{sim}(\cdot, \cdot)$ represents the cosine similarity. The alignment loss is defined as:
% Specifically, we align the projected hidden state feature \scalebox{0.9}{$h_\phi(H_t^{[n]})$} with the reference representation $F_0 = [cls_0; f_0] \in \mathbb{R}^{(N+1) \times D_{vf}}$. 
% Here, \scalebox{0.85}{$H_t^{[n]} \in \mathbb{R}^{(N+1) \times D_{z}} = \mathcal{E}_\theta(x_t^{[n]})$} denotes the output of the $n$-th transformer block, and $h_\phi$ is a trainable \textrm{MLP} projection, and $\operatorname{sim}(\cdot, \cdot)$ denotes a cosine similarity function. The alignment loss is formulated as:
\begin{equation}
\mathcal{L}_{\mathrm{REPA}}(\theta, \phi):=-\mathbb{E}_{\mathbf{x}_t, \epsilon, t}\left[ \frac{1}{N}\sum_{n=1}^{N}\operatorname{sim}(\mathbf{y}_{*}, h_\phi(\mathbf{h}_t^{[n]}))\right].
\end{equation}
In addition to alignment, the training objective includes velocity prediction for both the noised image latents $\mathbf{z}_t$ and class token $\mathbf{cls}_t$. The prediction loss is formulated as:
\begin{equation}
\mathcal{L}_{\text {pred }}=\mathbb{E}_{\mathbf{x}_*,\boldsymbol{\epsilon},{t}}\left[\left\|\mathbf{v}\left(\mathbf{z}_t, t\right)-\dot{\alpha}_t \mathbf{z}_*-\dot{\sigma}_t \epsilon_z\right\|^2+\beta\left\|\mathbf{v}\left(\mathbf{cls}_t, t\right)-\dot{\alpha}_t \mathbf{cls}_*-\dot{\sigma}_t \epsilon_{c l s}\right\|^2\right],
\end{equation}
% \begin{equation}
% \mathbb{E}_{\mathbf{x}_*,\boldsymbol{\epsilon},{t}}  [\left\| \mathbf{v}_{\theta}(\mathbf{x}_t, t) - \dot{\alpha}_t \mathbf{x}_* - \dot{\sigma}_t \boldsymbol{\epsilon} \right\|^2] .
% \end{equation}
% \begin{equation}
%     \dot{\mathbf{x}}_t = \mathbf{v}(\mathbf{x}_t, t),
% \end{equation}

where $\mathbf{v}\left ( \cdot ,t \right ) $ is the velocity prediction function, and $\beta>0$ controls the relative weighting between the image latents and class token denoising objectives. 
The final training loss integrates both prediction and alignment objectives, where $\lambda>0$ governs the relative weight of the alignment loss compared to the denoising loss. Specifically, the total loss $\mathcal{L}_{\text {total }}$ is formulated as:

\begin{equation}
\mathcal{L}_{\text {total }}=\mathcal{L}_{\mathrm{pred }} + \lambda \mathcal{L}_{\mathrm{REPA}}.
\end{equation}

\textbf{REG inference process.} The framework requires no auxiliary networks to generate the class token. 
REG jointly reconstructs both image latents and corresponding global semantics from random noise initialization. It acquired semantic knowledge actively guides and enhances the generation quality. 

In general, REG demonstrates three key advantages over existing approaches:
%\textbf{(1) Full utilization of discriminative information.}
\textbf{(1) Improved utilization of discriminative information.}
REG directly integrates discriminative information as part of the input during training, enabling both autonomous generation and consistent application of semantic guidance during inference. 
% This addresses a critical limitation of REPA, which relies on the external alignment mechanism that is absent during denoising inference.
This design aims to address a limitation of REPA, which cannot autonomously generate discriminative representations to guide generation in inference. Because it relies on an external alignment mechanism during training to utilize discriminative features, rather than incorporating them as input and applying the corresponding denoising task.
\textbf{(2) Minimal computational overhead.}
This design introduces only a single global class token, providing efficient and effective discriminative guidance while incurring an almost negligible computational overhead of less than $0.5\%$ FLOPs and latency at 256$\times$256 resolution (see Tab.~\ref{tab:flops}). 
\textbf{(3) Enhanced performance across metrics.}
REG improves superior performance in generation fidelity, accelerating training convergence, and discriminative semantic learning.
As shown in Fig.~\ref{fig:angle_resolver}(e), REG achieves up to 23$\times$ and 63$\times$ faster FID convergence than REPA and SiT, significantly reducing training time. Fig.~\ref{fig:cknna-repa} further shows consistently higher CKNNA scores across training steps, network layers, and timesteps.

\section{Experiments}
\label{exp}
In this section, we investigate three key research questions to evaluate the effectiveness and scalability of REG through comprehensive experimentation:
\begin{itemize}[leftmargin=20pt]\setlength{\itemsep}{5pt}
    \item[$\bullet$] \textbf{Model performance.} Can REG simultaneously accelerate training convergence and enhance generation quality? (Sec.~\ref{4.2})
    \item[$\bullet$] \textbf{Ablation analysis.} Verify the effectiveness of the REG different designs and hyperparameters. (Sec.~\ref{4.3})
    \item[$\bullet$]  \textbf{Discriminative semantics.} 
    Can REG improve the discriminative semantics of generative models? (Sec.~\ref{4.4})
\end{itemize}

\begin{figure}[t!]
\vspace{0.05in}
\begin{minipage}{0.4\textwidth}
\centering\small
\captionof{table}{\textbf{FID comparison across training iterations for accelerated alignment methods.} All experiments are conducted on ImageNet 256$\times$256 without classifier-free guidance (CFG). }

\vspace{-0.1in}
\resizebox{1\textwidth}{!}{%
\begin{tabular}{lccc}
\toprule
     Method  & \#Params & Iter. & FID$\downarrow$  \\
     \midrule
     SiT-B/2  &130M & 400K & 33.0 \\
     {+ REPA} &130M & 400K  & 24.4\\
     {\textbf{+ REG (ours)}} &132M & \textbf{400K}  & \textbf{15.2}\\
     \arrayrulecolor{black!40}\midrule
     SiT-L/2 &458M & 400K & 18.8 \\
     {+ REPA} &458M & 400K  & 9.7\\
     {+ REPA} &458M & 700K  & 8.4\\
     {\textbf{+ REG (ours)}} &460M & \textbf{100K}  & \textbf{\pz11.4}\\
     {\textbf{+ REG (ours)}} &460M & \textbf{400K} & \textbf{\pz4.6} \\
     \arrayrulecolor{black!40}\midrule
     SiT-XL/2  &675M & 400K   & 17.2\\
     SiT-XL/2  &675M & 7M   & \pz8.3\\
     {+ REPA} &675M & 200K  & 11.1\\
     {+ REPA} &675M & 400K  & 7.9\\
     {+ REPA} &675M & 1M  & 6.4\\
     {+ REPA} &675M & 4M  & 5.9\\
     {\textbf{+ REG (ours)}}  &677M & \textbf{200K}  & \textbf{5.0\pz}\\
     {\textbf{+ REG (ours)}} &677M & \textbf{400K}  & \textbf{3.4\pz}\\
     {\textbf{+ REG (ours)}} &677M & \textbf{1M}  & \textbf{2.7\pz}\\
     {\textbf{+ REG (ours)}} &677M & \textbf{2.4M}  & \textbf{2.2\pz}\\
     {\textbf{+ REG (ours)}} &677M & \textbf{4M}  & \textbf{1.8\pz}\\
     \arrayrulecolor{black}\hline
\end{tabular}
\label{tab:wo_cfg}
}
\end{minipage}
~~
\begin{minipage}{0.54\textwidth}

\centering\large
\captionof{table}{\textbf{Comparison of the performance of different methods on ImageNet 256$\times$256 with CFG.} Performance metrics are annotated with $\uparrow $ (higher is better) and $\downarrow$ (lower is better). }
\vspace{-0.05in}
\resizebox{\textwidth}{!}{%
\begin{tabular}{l c c c c c c}
\toprule
{\pz\pz Method} & Epochs  &  {\pz FID$\downarrow$} & {sFID$\downarrow$} & {IS$\uparrow$} & {Pre.$\uparrow$} & Rec.$\uparrow$ \\
\arrayrulecolor{black}\midrule

\multicolumn{7}{l}{\emph{Pixel diffusion}\vspace{0.02in}} \\
\pz\pz {ADM-U~\cite{ADM}}  &\pz400 &  \pz3.94 & 6.14 &  186.7 & 0.82 & 0.52 \\
\pz\pz VDM$++$~\cite{vdm} & \pz560 & \pz2.40 & - &  225.3 & - & - \\
\pz\pz Simple diffusion~\cite{simple} & \pz800 & \pz2.77 & - & 211.8 & - & - \\
\pz\pz CDM~\cite{CDM}  & 2160 & \pz4.88 & - & 158.7 & - & - \\
\arrayrulecolor{black!40}\midrule

\multicolumn{7}{l}{\emph{Latent diffusion, U-Net}\vspace{0.02in}} \\
\pz\pz LDM-4~\cite{ldm} & \pz200  & \pz3.60 & - & 247.7 & {0.87} & 0.48 \\
\arrayrulecolor{black!40}\midrule

\multicolumn{7}{l}{\emph{Latent diffusion, Transformer + U-Net hybrid}\vspace{0.02in}} \\
\pz\pz U-ViT-H/2~\cite{U-VIT} & \pz240 & \pz2.29 & 5.68  & 263.9 & 0.82 & 0.57 \\ 
\pz\pz DiffiT~\cite{diffit} & - & \pz1.73 & - &  276.5 & 0.80 & 0.62 \\
\pz\pz MDTv2-XL/2~\cite{mdtv2} & 1080  &  \pz1.58 & 4.52 & 314.7  & 0.79 & {0.65}\\
\arrayrulecolor{black}\midrule

\multicolumn{7}{l}{\emph{Latent diffusion, Transformer}\vspace{0.02in}} \\
\pz\pz MaskDiT~\cite{maskdit} & 1600 &  \pz2.28 & 5.67 & 276.6 & 0.80 & 0.61 \\ 
\pz\pz SD-DiT~\cite{sd-dit} & \pz480 & \pz3.23 & -    & -     & -    & -     \\
\arrayrulecolor{black!30}\cmidrule(lr){1-7}
\pz\pz {DiT-XL/2}~\cite{dit}   & 1400  &    \pz2.27 & 4.60 & {278.2} & {\textbf{0.83}} & 0.57  \\
\arrayrulecolor{black!30}\cmidrule(lr){1-7}
\pz\pz {SiT-XL/2}~\cite{sit}    & 1400 &     \pz2.06 & {4.50} & 270.3 & 0.82 & 0.59 \\
{{\pz\pz{+ REPA}}} & {{\pz800}} &{\pz1.42} & {4.70} & {{305.7}} & {0.80} & {{0.65}} \\

{\pz\pz{\textbf{+ REG (ours)}}} & {\pz80} & {\pz1.86} & {{4.49}} & {{\textbf{321.4}}} & {0.76} & {0.63} \\
{\pz\pz{\textbf{+ REG (ours)}}} & {\pz160} & {\pz1.59} &{{4.36}} & {{304.6}} & {0.77} & {0.65} \\
{\pz\pz{\textbf{+ REG (ours)}}} & {\pz\textbf{480}} &1.40  &\textbf{4.24}  &296.9  &0.77  &\textbf{0.66} \\
{\pz\pz{\textbf{+ REG (ours)}}} & {\pz\textbf{800}} &\textbf{1.36}  &4.25  &299.4  &0.77  &\textbf{0.66} \\
\arrayrulecolor{black}\bottomrule
\end{tabular}
}
\label{tab:main}
\end{minipage}
\vspace{-0.15in}
\end{figure}

\subsection{Setup}
\label{setup}

\textbf{Implementation details.} 

We adhere strictly to the standard training protocols of SiT~\cite{sit} and REPA~\cite{repa}. Experiments are conducted on the ImageNet dataset~\cite{imagenet}, with all images preprocessed to 256$\times$256 resolution via center cropping and resizing, following the ADM framework~\cite{ADM}. Each image is encoded into a latent representation $z \in \mathbb{R}^{32 \times 32 \times 4}$ using the Stable Diffusion VAE~\cite{ldm}. Model architectures B/2, L/2, and XL/2 (with $2\times2$ patch processing) follow the SiT specifications~\cite{sit}. For comparability, we fix the training batch size to 256 and adopt identical learning rates and Exponential Moving Average (EMA) configurations as REPA~\cite{repa}. Additional implementation details are provided in the \textit{Appendix}.

\textbf{Evaluation protocol.} 
 
To comprehensively evaluate image generation quality across multiple dimensions, we employ a rigorous set of quantitative metrics including Fréchet Inception Distance (FID)~\cite{fid} for assessing realism, structural FID (sFID)~\cite{sfid} for evaluating spatial coherence, Inception Score (IS)~\cite{is} for measuring class-conditional diversity, precision (Prec.) for quantifying sample fidelity, and recall (Rec.)~\cite{rec} for evaluating coverage of the target distribution, all computed on a standardized set of 50K generated samples to ensure statistical reliability. We further supplement these assessments with CKNNA~\cite{cknna} for analyzing feature-space characteristics. Sampling follows REPA~\cite{repa}, using the SDE Euler–Maruyama solver with 250 steps. Full evaluation protocol details are provided in the~\textit{Appendix}.

\begin{table}[t]
  \centering
  \small
  \setlength{\tabcolsep}{10pt}
   \caption{\textbf{Verify the effects of various target representation (Target Repr.)~\cite{dinov2, clip}, the depth of supervision (Depth), and the loss weight ($\beta$)}. Experiments employ SiT-B/2 architectures trained for 400K iterations on ImageNet 256$\times$256. Performance metrics (with ↓/↑ denoting preferred directions) are computed using an SDE Euler-Maruyama sampler (NFE=250) without classifier-free guidance. REPA$\dagger$ indicates our local reproduction of the original method's reported results.}
   \vspace{5pt}
  \resizebox{1\linewidth}{!}
    {
      \begin{tabular}{l|c|c|c|ccccc}
        \hline
        Method & Target Repr.  &Depth &$\beta$ & FID$\downarrow$ &  sFID$\downarrow$ & IS$\uparrow$ & Pre.$\uparrow$ &Rec.$\uparrow$\\
        \hline
        SiT-B/2   &-   &-  &- &33.00 &6.46 &43.70 &0.53 &0.63   \\
        + REPA    &DINOv2-B  &4  &-   & 24.40 &6.40 &59.90 &0.59 &0.65  \\
        + REPA$\dagger$   &DINOv2-B  &4 &-  & 22.38 &6.98 &66.65 &0.59 &0.65 \\
        \hline
        \multirow{12}{*}{+ REG} &\cellcolor{tabhighlight}CLIP-L     &4 &0.03 &21.30 &6.51 &70.14 &0.61 &0.63    \\
        &\cellcolor{tabhighlight}DINOv2-B   &4 &0.03 &\cellcolor{tabhighlight}15.22 &\cellcolor{tabhighlight}6.89 &\cellcolor{tabhighlight}94.64 &\cellcolor{tabhighlight}0.64 &\cellcolor{tabhighlight}0.63  \\
        &\cellcolor{tabhighlight}DINOv2-L   &4 &0.03 &17.36 &7.02 &89.88 &0.63 &0.63 \\
        \cline{2-9}
        &DINOv2-B   &\cellcolor{tabhighlight}2 &0.03 &18.19 &6.67 &83.96 &0.62 &0.64   \\
        &DINOv2-B   &\cellcolor{tabhighlight}4 &0.03 & \cellcolor{tabhighlight}15.22 &\cellcolor{tabhighlight}6.89 &\cellcolor{tabhighlight}94.64 &\cellcolor{tabhighlight}0.64 &\cellcolor{tabhighlight}0.63   \\
        &DINOv2-B   &\cellcolor{tabhighlight}6 &0.03 & 16.31 &7.11 &91.72 &0.63 &0.64 \\
        &DINOv2-B   &\cellcolor{tabhighlight}8 &0.03 &17.31  &7.23  &87.78  &0.63 &0.63  \\
        \cline{2-9}
        &DINOv2-B   &4 &\cellcolor{tabhighlight}0.01  &15.76  &6.69  &93.75  &0.66 &0.61   \\
        &DINOv2-B   &4 &\cellcolor{tabhighlight}0.02 &15.64  &6.70  &93.86  &0.66 &0.63   \\
        &DINOv2-B   &4 &\cellcolor{tabhighlight}0.03 &\cellcolor{tabhighlight}15.22  &\cellcolor{tabhighlight}6.69  &\cellcolor{tabhighlight}94.64  &\cellcolor{tabhighlight}0.64 &\cellcolor{tabhighlight}0.63 \\
        &DINOv2-B   &4 &\cellcolor{tabhighlight}0.05 &16.28  &6.97  &92.39  &0.64 &0.64  \\
        &DINOv2-B   &4 &\cellcolor{tabhighlight}0.10 &18.41  &7.44  &84.79  &0.61 &0.64  \\
        \hline
      \end{tabular}
    }
    %\vspace{-1.5em}
    \label{tab:cims}
    %\vspace{-5pt}
\end{table}

\subsection{Improving the performance of generative models}
\label{4.2}

\textbf{Accelerating training convergence.} Tab.~\ref{tab:wo_cfg} provides a detailed comparison between REG, SiT~\cite{sit}, and REPA~\cite{repa} across multiple model scales on ImageNet 256$\times$256 without CFG. 
The proposed REG framework consistently achieves the lowest FID scores while substantially accelerating training across all configurations.
\textbf{For smaller models}, SiT-B/2 + REG outperforms SiT-B/2 + REPA by 9.2 FID points and surpasses SiT-L/2 trained for 400K iterations by 3.6 points. 
\textbf{In the large-scale models}, SiT-L/2 + REG achieves an FID of 4.6 at 400K steps, outperforming both SiT-XL/2 + REPA (4M steps) by 1.3 points and SiT-XL/2 (7M steps) by 3.7 points, while requiring only 10.0\% and 5.71\% of their respective training costs.
Similarly, SiT-XL/2 + REG achieves comparable performance to SiT-XL/2 (7M steps) and REPA-XL/2 (4M steps) in just 110K and 170K steps, respectively, demonstrating 63$\times$ and 23$\times$ faster convergence (see Fig.~\ref{fig:angle_resolver}(e)).
At 4M steps, REG achieves a record-low FID of 1.8, demonstrating superior scalability and efficiency across model sizes.

\textbf{Comparison with SOTA methods.} Tab.~\ref{tab:main} presents a comprehensive comparison against recent SOTA methods utilizing classifier-free guidance. 
Our framework achieves competitive performance using the REPA's same guidance interval~\cite{kynkaanniemi2024applying} with significantly reduced training cost.
REG matches SiT-XL's quality in just 80 epochs (17$\times$ faster than SiT-XL's 1400 epochs) and surpasses REPA's 800-epoch performance at 480 epochs, highlighting its superior training efficiency and convergence properties. Additional experiments in the \textit{Appendix} include the results of more training steps, validating the REG's robustness, scalability, and cross-task generalization.

\begin{wraptable}{rt}{0.6\linewidth}
    \vspace{-1em}
    \centering
    \captionsetup{font={footnotesize}}
    \caption{\textbf{Computational cost and performance comparison.} This table compares REPA and REG on ImageNet 256$\times$256, detailing model size, FLOPs, sampling steps, latency, and generation quality metrics. REG achieves substantially better sample quality with negligible increases in computational cost.}
    \vspace{-3pt}
    \scriptsize
    \setlength{\tabcolsep}{2pt}
    \resizebox{0.99\linewidth}{!}
    {
      \begin{tabular}{l|ccc|cc}
        \hline
        Method &\#Params & FLOPs$\downarrow$  &Latency (s)$\downarrow$  & FID$\downarrow$  & IS$\uparrow$ \\
        \hline
        SiT-XL/2 + REPA  &675 &114.46   & 6.18 &7.90 &122.60   \\
        \hline
        \multirow{2}{*}{SiT-XL/2 + REG}  &\multirow{2}{*}{{\shortstack[c]{677\\ (+0.30\%)}}} &\multirow{2}{*}{{\shortstack[c]{114.90\\ (+0.38\%)}}}  &\multirow{2}{*}{{\shortstack[c]{6.21\\ (+0.49\%)}}} &\multirow{2}{*}{{\shortstack[c]{3.44\\ (+56.46\%)}}}  &\multirow{2}{*}{{\shortstack[c]{184.13\\ (+50.19\%)}}}  \\
        &&&&\\
        \hline
      \end{tabular}
    }
    \label{tab:flops}
    \vspace{-1em}
\end{wraptable}

\textbf{Computational cost comparison.} We compare the computational efficiency of REG and REPA under the same model scale (SiT-XL/2) in Tab.~\ref{tab:flops}. REG introduces only a marginal increase in parameter count (+0.30\%) and FLOPs (+0.38\%) relative to REPA, while maintaining nearly identical latency (6.21s vs. 6.18s, +0.49\%). 
Despite the minimal computational overhead, REG yields substantial improvements in generation quality, achieving a 56.46\% relative reduction in FID, alongside a 50.19\% increase in IS. These results demonstrate that REG simultaneously improves generation quality and computational efficiency, highlighting its effectiveness as a general-purpose enhancement for generative models.

\subsection{Ablation Studies}
\label{4.3}

\textbf{Different discriminative guidance.} We systematically investigate the impact of different pretrained vision encoders and their corresponding class tokens as target representations in Tab.~\ref{tab:cims}. 
Among all configurations, DINOv2-B achieves the best performance with the lowest FID (15.22) and highest IS (94.64). Notably, all evaluated target representations consistently surpass the REPA, providing empirical evidence that class tokens derived from self-supervised models enhance generation fidelity.

\textbf{Alignment depth.} As shown in Tab.~\ref{tab:cims}, we compare the effects of applying the REPA loss at different network depths. Our analysis reveals that applying the loss in earlier layers yields superior results, which is consistent with REPA's findings. Notably, our method demonstrates consistent improvements over REPA across all configurations, achieving FID reductions ranging from 4.19 to 7.16 points.
We attribute these gains to the direct insertion of the class token, which provides discrete global guidance to all layers. This enables adaptive integration of discriminative semantics throughout the network, in contrast to REPA’s indirect supervision mechanism, where only selected features are aligned with the target representation. As a result, REG allows remaining layers to capture richer high-frequency details than REPA, contributing to the observed improvements.

\textbf{Effect of $\beta$.} Tab.~\ref{tab:cims} systematically evaluates the impact of varying the loss weight $\beta$, which controls the contribution of the class token alignment loss. Among the tested values, $\beta = 0.03$ achieves the best overall performance across all evaluation metrics. Consequently, this value was adopted as the default parameter for all subsequent experiments.

\begin{wraptable}{rt}{0.5\linewidth}
    \vspace{-1.2em}
    \centering
    \captionsetup{font={footnotesize}}
    \setlength{\tabcolsep}{3pt}
    \caption{\textbf{Ablation study on different entanglement signals.} All experiments are conducted on ImageNet 256$\times$256, using SiT-B/2 models trained for 400K iterations. This experiment adopts the best configuration from Tab.~\ref{tab:cims} and focuses solely on the impact of different entanglement signals on generation quality.}
    \vspace{-3pt}
    \scriptsize
    \resizebox{0.95\linewidth}{!}{
    \begin{tabular}{l|c c c  }
        \hline
        Method  &FID$\downarrow$ & sFID$\downarrow$ & IS$\uparrow$  \\
        \hline
        SiT-B/2 + REPA & 24.40 & 6.40 & 59.90  \\
        \hline
        + one learnable token   &23.31   &6.48   &63.44     \\
        + avg (latent features)  &24.12   &6.52   &60.78      \\
        + avg (DINOv2 features)  &16.86   &6.67   &84.91     \\
        + \textbf{DINOv2 class token}  &\textbf{15.22}   &\textbf{6.69}   &\textbf{94.64}    \\
        \hline
    \end{tabular}
    }
    \vspace{-1em}
    \label{tab:different_size1}
\end{wraptable}

\textbf{Entanglement signal variants.} Tab.~\ref{tab:different_size1} systematically evaluates the impact of different entanglement signals on generation quality through concatenative operation. 
Concatenating noised latent features with either a learnable token or the average of latent features provides limited improvement, likely due to the lack of rich discriminative semantic information
In contrast, incorporating discriminative signals yields substantial gains: averaged DINOv2 features significantly reduce FID to 16.86, while the DINOv2 class token achieves the best performance, lowering FID by 9.18 and increasing IS to 94.64. 
These results yield two key insights: (1) high-level discriminative information (class token) substantially enhances generation quality, and (2) the entanglement methodology critically governs performance improvements. 
The demonstrated efficacy of class token concatenation reveals that global discriminative information effectively regularizes the generative latent space, simultaneously boosting both semantic and output quality while maintaining computational efficiency.

\begin{wraptable}{rt}{0.5\linewidth}
    \vspace{-1.2em}
    \centering
    \setlength{\tabcolsep}{3pt}
    \captionsetup{font={footnotesize}}
    \caption{\textbf{Ablation study on different class token entanglement~\cite{dinov2, clip} without representation alignment.} This table investigates the effectiveness of class token entanglement in the absence of explicit representation alignment. All experiments are conducted on ImageNet 256$\times$256 at 400K iterations.}
    \vspace{-3pt}
    \scriptsize
    \resizebox{0.99\linewidth}{!}{
    \begin{tabular}{l|c|ccc}
        \hline
        Method & Class token &FID$\downarrow$ & sFID$\downarrow$ & IS$\uparrow$  \\
        \hline
        SiT-B/2 &- &33.0 &6.46 &43.70  \\
        \hline
        \multirow{3}{*}{+ Entanglement} & CLIP-L    &32.05   &6.76   &47.61     \\
        &\textbf{DINOv2-B}  &\textbf{26.67}   &\textbf{6.88}   &\textbf{59.37}     \\
        &DINOv2-L  &30.16   &6.91   &52.86      \\
        \hline
    \end{tabular}
    }
    \label{tab:different_size2}
    \vspace{-1em}
\end{wraptable}

\textbf{Effectiveness of entanglement alone.} Tab.~\ref{tab:different_size2} evaluates the impact of incorporating class tokens from various pretrained self-supervised encoders into SiT-B/2 without applying representation alignment. 
The results demonstrate that class token entanglement alone consistently enhances generation quality, with FID improvements ranging from 0.95 to 6.33 points across all variants. 
Notably, DINOv2-B delivers optimal performance, achieving a 19.18\% FID reduction and 35.86\% IS improvement compared to the SiT-B/2 baseline. 
These findings indicate that the model can effectively leverage high-level semantic guidance from the class token, even in the absence of explicit alignment, highlighting the robustness and general utility of class token-based entanglement for generative modeling.

\vspace{-1em}
\subsection{Enhancing the discriminative semantic learning of generative models}
\label{4.4}

\begin{figure}[t]
    \centering
    \includegraphics[width=1.0\linewidth]{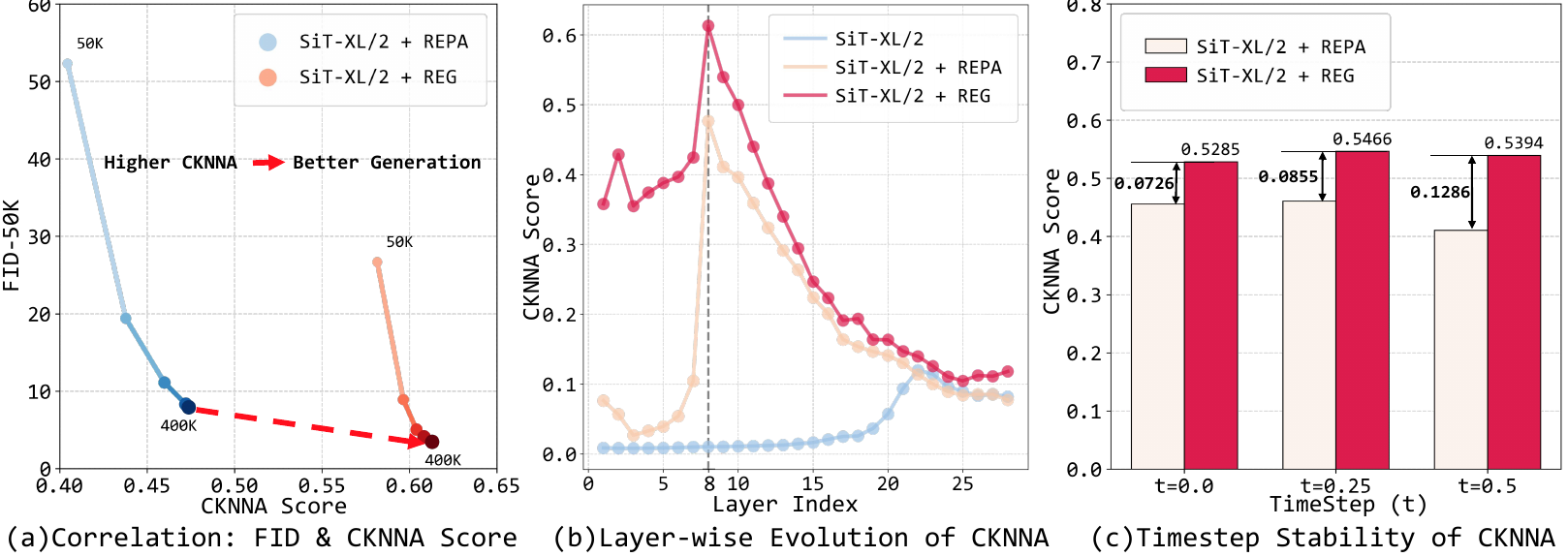}
    \caption{\textbf{Analysis of Discriminative Semantics.}
    (a) Correlation between CKNNA and FID across training steps (color-coded by progression). REG demonstrates superior discriminative semantic learning, achieving higher CKNNA scores alongside lower FID values compared to REPA.
    (b) Layer-wise CKNNA progression at 400K training steps (t=0.5). REG consistently enhances CKNNA across all network layers, indicating robust discriminative semantics learning.
    (c) Timestep-wise CKNNA variation at 400K training. REG improves semantic alignment uniformly throughout the training process, outperforming baselines at all timesteps.
    }
    \label{fig:cknna-repa}
\end{figure}

We systematically measure REG, SiT, and REPA's CKNNA scores across training steps, network layers, and timesteps to assess the discriminative semantics of dense features. For fair comparison, we follow REPA’s evaluation protocol: We compute CKNNA scores exclusively between spatially averaged generative model dense features and averaged DINOv2-g dense features, while class token are not involved in calculations. Here are the specific situations:

\textbf{Training steps analysis.} Fig.~\ref{fig:cknna-repa}(a) shows the positive correlation between CKNNA and FID scores across training steps at layer 8 (t=0.5). It reveals that both REPA and REG achieve improved semantic alignment (higher CKNNA) with better generation quality (lower FID). Notably, REG consistently outperforms REPA in both metrics throughout training, demonstrating its superior capacity for discriminative semantic learning through discriminative semantics guidance.

\textbf{Layer-wise progression.} At 400K training steps (t=0.5) in Fig.~\ref{fig:cknna-repa}(b), both REG and REPA exhibit similar CKNNA patterns: semantic scores gradually increase until reaching the peak at layer n=8 (where alignment loss is computed), then progressively decrease. Crucially, REG achieves consistently higher semantic scores than REPA and SiT across all network layers. This improvement stems from REG's innovation of entangling low-level image latents with high-level class token from pretrained foundation models. Through attention mechanisms, REG effectively propagates these discriminative semantics to guide the model in understanding low-level features in early layers, while later layers subsequently focus on predicting high-frequency details.

\textbf{Timestep robustness.} Evaluation of CKNNA at layer 8 (400K steps) demonstrates REG's consistent superiority across all timesteps in Fig.~\ref{fig:cknna-repa}(c). This robustness confirms its stable, high-level semantic guidance capability throughout the entire noise spectrum, enabling reliable discriminative semantic performance regardless of noise intensity during generation.

\section{Conclusion}
This paper presents Representation Entanglement for Generation (REG), a simple and efficient framework that firstly introduces image-class denoising paradigm instead of the current pure image denoising pipeline, which fully unleashes the potential of discriminative gains for generation.
REG entangles low-level image latents with a single high-level class token from pretrained foundation models, achieved via synchronized noise injection and spatial concatenation. 
The denoising process simultaneously reconstructs both image latents and corresponding global semantics, enabling active semantic guidance that enhances generation quality while introducing minimal computational cost through the addition of just one token.
Extensive experiments demonstrate REG's superior performance in generation fidelity, accelerating training convergence, and discriminative semantic learning, validating its effectiveness and scalability.

\newpage
%\printbibliography
\bibliographystyle{unsrt}
\bibliography{Reference}
%\small
%\bibliography{Reference}
%\normalsize
\newpage
\maketitlesupplementary
\appendix

\section{Discriminative semantics in inference}

\begin{figure*}[h]
    \centering
    \includegraphics[width=0.85\linewidth]{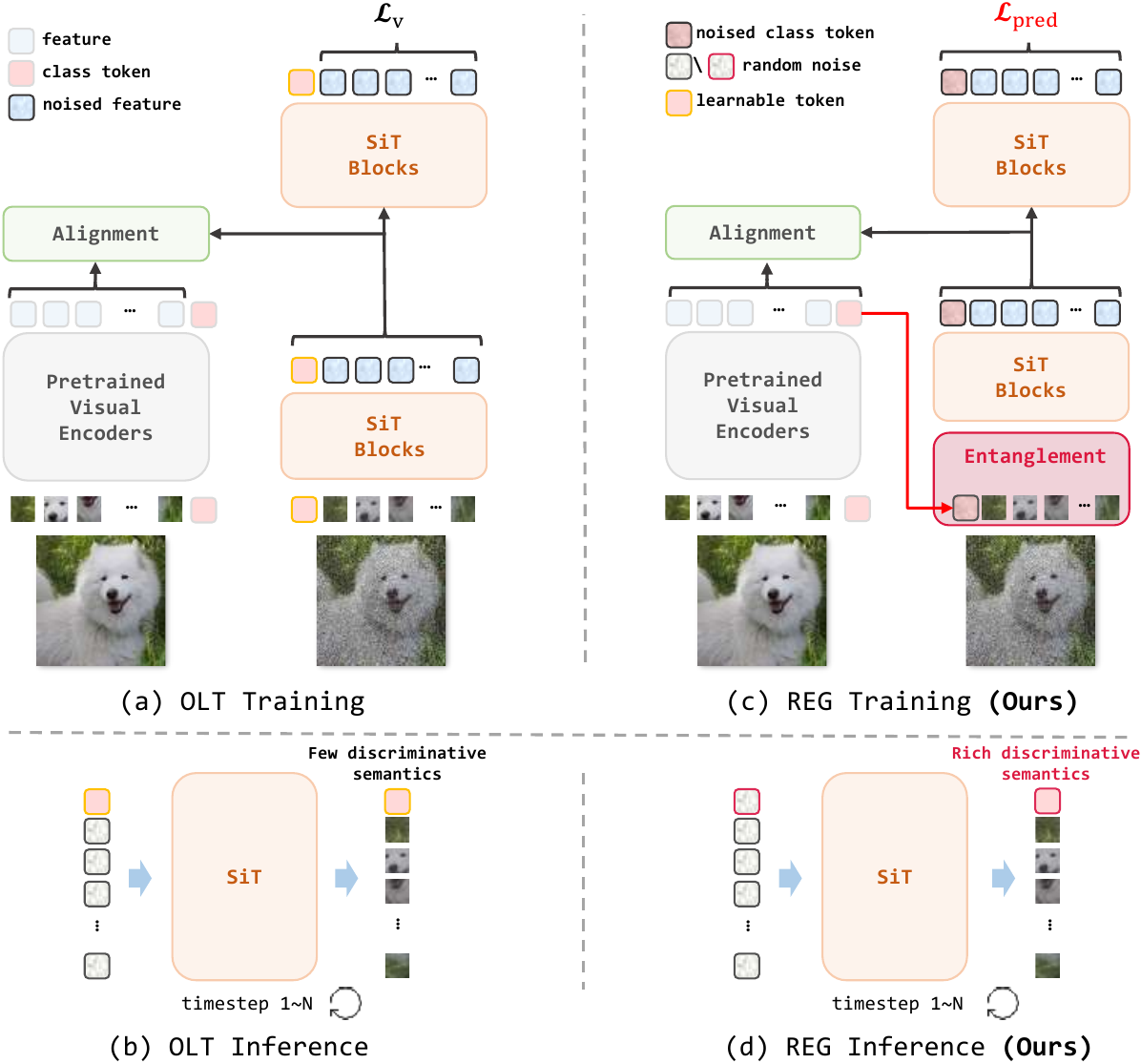}
    \caption
    {
        Comparison between the One Learnable Token (OLT) method and REG for generation.  
        (a) During training, OLT simply concatenates noised latents with a learnable token while computing velocity prediction loss~$\mathcal{L}_{\mathbf{v}}$ on dense features.
        (b) OLT's output learnable token demonstrates minimal discriminative semantics after multi-step denoising.
        (c) REG entangles one high-level noised class token with low-level noised latent features while computing velocity prediction losses~$\mathcal{L}_{\text {pred}}$ for both components. 
        (d) REG can reconstruct the corresponding global semantics of image latents with rich discriminative semantics.
    }
    \label{fig:app_com}
\end{figure*}

We posit that REG's semantic reconstruction capability during inference stems from two key design elements: (1) the architectural entanglement of class token with image latents during training, and (2) the consistent application of SiT's~\cite{sit} velocity prediction loss to both them. 
Comparative analysis reveals: (1) The One learnable Token (OLT) method (see Fig.~\ref{fig:app_com}(a)) concatenates noised latents with a learnable token and only calculates velocity prediction loss~$\mathcal{L}_{\mathbf{v}}$ on dense features. In contrast, REG (see Fig.~\ref{fig:app_com}(c)) entangles one high-level noised class token with low-level noised latent features while computing velocity prediction losses~$\mathcal{L}_{\text {pred }}$ for both components.

\begin{figure*}[h]
    \vspace{-10pt}
    \centering
    \includegraphics[width=0.45\linewidth]{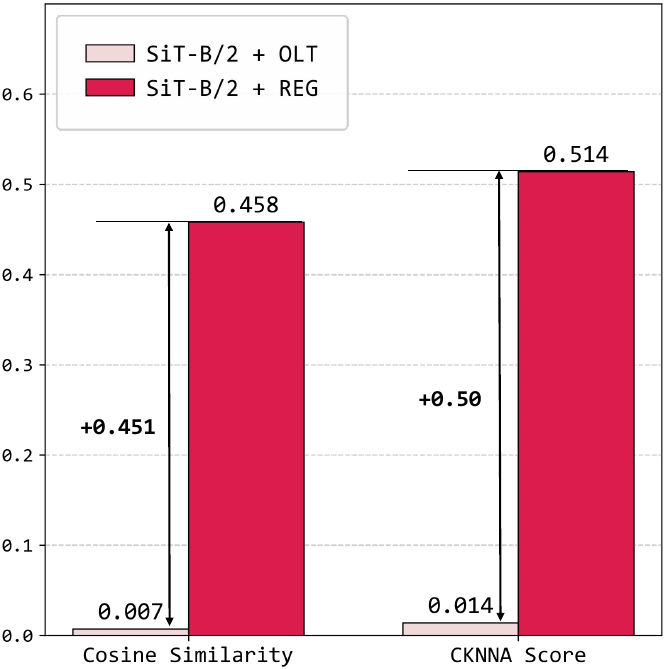}
    %\captionsetup{font={footnotesize}}
    \caption
    {
        Quantitative evaluation of cosine similarity and CKNNA score between OLT's learnable token, REG's class token, and DINOv2-g's reference token. Results demonstrate REG's superior semantic retention during inference compared to OLT.
    }
    \label{fig:c2}
    \vspace{-15pt}
\end{figure*}

To quantitatively validate the two methods' discriminative semantics in inference, we use 10,000 ImageNet validation images as input processed through identical noise injection via the VAE encoder~\cite{ldm}. 
REG's inference integrates noised latents with the noise-initialized class token through concatenation before multi-step denoising (see Fig.~\ref{fig:app_com}(d)), while OLT similarly processes noised latents with its learnable token (see Fig.~\ref{fig:app_com}(b)). 
Then, we process these ImageNet validation images through DINOv2~\cite{dinov2} to obtain the reference class token. Fig.~\ref{fig:c2} computes both CKNNA and cosine similarity between the reference class token and the output from REG's class token/OLT's learnable token using SiT-B/2 as backbone. 
The results demonstrate a significant disparity: OLT's learnable token achieves only 0.007 cosine similarity and 0.014 CKNNA scores, while REG's class token reaches 0.458 and 0.514, respectively. 
This empirical evidence confirms REG's superior capacity to preserve discriminative semantics compared to OLT's limited representation capability in inference.

\section{Analysis of training overhead in REG}
We summarize the total training overhead in Tab.~\ref{tab:training_gpus}, reporting the costs required to reach the same performance upper bounds claimed in the original SiT and REPA papers. All experiments are conducted on 8 NVIDIA A40 GPUs. Our results show that REG requires only 110K training steps to reach the performance level of SiT trained for 7M steps, reducing GPU hours by 98.36\%. 
Moreover, compared with REPA, REG achieves the performance of its 4M counterpart with only 170K iterations, reducing GPU hours by 95.72\%.
These results highlight the training efficiency of REG, demonstrating faster convergence and significantly lower training overhead compared to prior methods.

\begin{table}[ht!]
\centering\small
\setlength{\tabcolsep}{3pt}
\caption{Training overhead comparison. REG achieves comparable performance to other models while significantly reducing training time.}
\label{tab:training_gpus}
\resizebox{0.5 \textwidth}{!}{%\resizebox{0.95\linewidth}{!}
\begin{tabular}{lccccccccc}
\toprule
     Model  & FID$\downarrow$ & Training step$\downarrow$  & All GPU hours$\downarrow$ \\
     \midrule
     \rowcolor{gray!15}
     SiT-XL/2 & 8.3  & 7M   &2380  \\
     \textbf{{+ \sname (ours)}} & \textbf{8.2} & \textbf{110K} & \textbf{39 (-98.36\%)}	  \\
     \arrayrulecolor{black!40}\midrule
     {+ REPA} & 5.9 & 4M & 1800	  \\
     \textbf{{+ \sname (ours)}} & \textbf{5.6} & \textbf{170K} & \textbf{77 (-95.72\%)}	  \\
     \arrayrulecolor{black}\hline
\end{tabular}
}
\end{table}

\section{256$\times$256 ImageNet}

Tab.~\ref{tab:256_sota} presents extended training results with CFG using the REPA's same guidance interval~\cite{kynkaanniemi2024applying}, demonstrating REG's excellent performance with a 1.40 FID at 480 epochs; it achieves better performance comparable to REPA~\cite{repa} at 800 epochs while requiring fewer than 40\% of the training iterations.
In addition, Tab.~\ref{tab:detail} presents more specific performance details of SiT + REG, further highlighting its superior robustness and accelerated convergence. Tab.~\ref{tab:cfg_interval} presents quantitative performance metrics of SiT-XL + REG under varying classifier-free guidance scale $w$.

\begin{table}[h]
  \centering
  \small
  \vspace{-8pt}
  \caption{Extended REG training on ImageNet 256$\times$256 with CFG demonstrates progressive performance gains.}
  \setlength{\tabcolsep}{12pt}
  \resizebox{0.9\linewidth}{!}
    {
      \begin{tabular}{lcccccc}
        \toprule
        {\pz\pz Method} & Epochs  &  {\pz FID$\downarrow$} & {sFID$\downarrow$} & {IS$\uparrow$} & {Pre.$\uparrow$} & Rec.$\uparrow$ \\
        \arrayrulecolor{black}\midrule
        
        \multicolumn{7}{l}{\emph{Pixel diffusion}\vspace{0.02in}} \\
        \pz\pz {ADM-U~\cite{ADM}}  &\pz400 &  \pz3.94 & 6.14 &  186.7 & 0.82 & 0.52 \\
        \pz\pz VDM$++$~\cite{vdm} & \pz560 & \pz2.40 & - &  225.3 & - & - \\
        \pz\pz Simple diffusion~\cite{simple} & \pz800 & \pz2.77 & - & 211.8 & - & - \\
        \pz\pz CDM~\cite{CDM}  & 2160 & \pz4.88 & - & 158.7 & - & - \\
        \arrayrulecolor{black!40}\midrule
        
        \multicolumn{7}{l}{\emph{Latent diffusion, U-Net}\vspace{0.02in}} \\
        \pz\pz LDM-4~\cite{ldm} & \pz200  & \pz3.60 & - & 247.7 & {0.87} & 0.48 \\
        \arrayrulecolor{black!40}\midrule
        
        \multicolumn{7}{l}{\emph{Latent diffusion, Transformer + U-Net hybrid}\vspace{0.02in}} \\
        \pz\pz U-ViT-H/2~\cite{U-VIT} & \pz240 & \pz2.29 & 5.68  & 263.9 & 0.82 & 0.57 \\ 
        \pz\pz DiffiT*~\cite{diffit} & - & \pz1.73 & - &  276.5 & 0.80 & 0.62 \\
        \pz\pz MDTv2-XL/2*~\cite{mdtv2} & 1080  &  \pz1.58 & 4.52 & 314.7  & 0.79 & {0.65}\\
        \arrayrulecolor{black}\midrule
        
        \multicolumn{7}{l}{\emph{Latent diffusion, Transformer}\vspace{0.02in}} \\
        \pz\pz MaskDiT~\cite{maskdit} & 1600 &  \pz2.28 & 5.67 & 276.6 & 0.80 & 0.61 \\ 
        \pz\pz SD-DiT~\cite{sd-dit} & \pz480 & \pz3.23 & -    & -     & -    & -     \\
        \arrayrulecolor{black!30}\cmidrule(lr){1-7}
        \pz\pz {DiT-XL/2}~\cite{dit}   & 1400  &    \pz2.27 & 4.60 & {278.2} & {0.83} & 0.57  \\
        \arrayrulecolor{black!30}\cmidrule(lr){1-7}
        \pz\pz {SiT-XL/2}~\cite{sit}    & 1400 &     \pz2.06 & {4.50} & 270.3 & 0.82 & 0.59 \\
        {{\pz\pz{+ REPA}}} & {{\pz800}} &{\pz1.42} & {4.70} & {{305.7}} & {0.80} & {{0.65}} \\
        
        {\pz\pz{\textbf{+ REG (ours)}}} & {\pz80} & {\pz1.86} & {{4.49}} & {{321.4}} & {0.76} & {0.63} \\
        {\pz\pz{\textbf{+ REG (ours)}}} & {\pz160} & {\pz1.59} &{{4.36}} & {{304.6}} & {0.77} & {0.65} \\
        {\pz\pz{\textbf{+ REG (ours)}}} & {\pz300} &{{1.48}} & {{4.31}} & {305.8} & {0.77} & {0.66} \\
        {\pz\pz{\textbf{+ REG (ours)}}} & {\pz480} &1.40  &4.24  &296.9  &0.77  &0.66\\
        {\pz\pz{\textbf{+ REG (ours)}}} & {\pz800} &1.36  &4.25  &299.4  &0.77  &0.66\\
        \arrayrulecolor{black}\bottomrule
    \end{tabular}
}
    \vspace{-10pt}
    \label{tab:256_sota}
       % \vspace{-5pt}
\end{table}

\begin{table}[h]
\centering\small
\caption{More performance analysis of SiT + REG across model scales without CFG.}
\begin{tabular}{lccccccc}
\toprule
     Model & \#Params & Iter. & FID$\downarrow$ & sFID$\downarrow$ & IS$\uparrow$ & Prec.$\uparrow$ & Rec.$\uparrow$  \\
     \midrule
     \rowcolor{gray!15}
     SiT-B/2 \citep{sit} & 130M & 400K 
     & 33.0 & \pz6.46 & \pz43.7 & 0.53 & 0.63  \\
     {{+ REPA}} & 130M & {400K} & 24.4 & \pz6.40 & \pz59.9 & 0.59 & 0.65 \\
     %\hline
     {{+ \sname (ours)}} & 132M & {50K}  &64.7  &9.47  &23.2  &0.40  &0.51 \\
     {{+ \sname (ours)}} & 132M & {100K} &36.1  &7.74  &45.5  &0.53  &0.61  \\
     {{+ \sname (ours)}} & 132M & {200K} &22.1  &7.19  &72.2  &0.60  &0.63  \\
     {{+ \sname (ours)}} & 132M & {400K} &15.2  &6.69  &94.6  &0.64  &0.63  \\
     \midrule
     \rowcolor{gray!15}
     SiT-L/2 \citep{sit} & 458M & 400K 
     & 18.8 & \pz5.29 & \pz72.0 & 0.64 & 0.64  \\
     {{+ REPA }} & 458M & {400K} & 10.0 & \pz5.20 & 109.2 & 0.69 & 0.65 \\  
     %\hline
     {{+ \sname (ours)}} & 460M & {50K}  &30.1  &8.92  &52.6  &0.58  &0.57  \\
     {{+ \sname (ours)}} & 460M & {100K} &11.4  &5.36  &108.8  &0.70  &0.60    \\
     {{+ \sname (ours)}} & 460M & {200K} &6.6  &5.16  &145.4  &0.73  &0.63  \\
     {{+ \sname (ours)}} & 460M & {400K} &4.6  &5.21  &167.6  &0.75  &0.63  \\
     \arrayrulecolor{black}\midrule
     \rowcolor{gray!15}
     SiT-XL/2 \citep{sit} & 675M & 7M   & \pz8.3 & \pz6.32 & 131.7 & 0.68 & 0.67 \\
     {{+ REPA}} & 675M & {4M}  & \pz5.9 & \pz5.73 & 157.8 & 0.70 & 0.69  \\
     %\hline
     {{+ \sname (ours)}} & 677M & {50K}  &26.7  &16.49  &59.2  &0.60  & 0.54  \\
     {{+ \sname (ours)}} & 677M & {100K} &8.9  &5.50  &125.3  &0.72  &0.59  \\
     {{+ \sname (ours)}} & 677M & {200K} &5.0  &4.88  &161.2  &0.75  &0.62  \\
     {{+ \sname (ours)}} & 677M & {400K} &3.4  &4.87  &184.1  &0.76  &0.64  \\
     {{+ \sname (ours)}} & 677M & {1M} &2.7  &4.93  &201.8  &0.76  &0.66  \\
     {{+ \sname (ours)}} & 677M & {2.4M} &2.2  &4.79  &219.1  &0.76  &0.66  \\
     {{+ \sname (ours)}} & 677M & {4M} &1.8  &4.59  &230.8  &0.77  &0.66  \\
\bottomrule
\end{tabular}
\label{tab:detail}
\vspace{-5pt}
\end{table}

\begin{table}[ht!]
\centering\small
\caption{The results of SiT-XL + REG at 2.4M training iterations under varying classifier-free guidance scale $w$, employing the guidance interval method~\cite{kynkaanniemi2024applying}.}
\label{tab:cfg_interval}
\resizebox{\textwidth}{!}{%
\begin{tabular}{lccccccccc}
\toprule
     Model & \#Params & Iter. & Interval & $w$  & FID$\downarrow$ & sFID$\downarrow$ & IS$\uparrow$ & Prec.$\uparrow$ & Rec.$\uparrow$  \\
     \midrule
     \rowcolor{gray!15}
     SiT-XL/2 \citep{sit} & 675M  & 7M &[0, 1] & 1.50  & 2.06 & 4.50 & 270.3 & 0.82 & 0.59  \\
     {{+ \sname (ours)}} & 675M & {2.4M} & [0, 0.8] &2.4   &1.45  &4.32  &280.44  &0.77  &0.67  \\
     {{+ \sname (ours)}} & 675M & {2.4M} & [0, 0.85] &2.4  &1.41  &4.24  &299.65  &0.77  &0.67 \\
     {{+ \sname (ours)}} & 675M & {2.4M} & [0, 0.9] &2.4   &1.61 &4.21  &334.50  &0.79  &0.64 \\
     \hline
     {{+ \sname (ours)}} & 675M & {2.4M} & [0, 0.85] &2.5   &1.43	&4.25	&303.11	 &0.77	&0.67 \\
     {{+ \sname (ours)}} & 675M & {2.4M} & [0, 0.85] &2.4   &1.41  &4.24  &299.65  &0.77  &0.67 \\
     {{+ \sname (ours)}} & 675M & {2.4M} & [0, 0.85] &2.3  &1.40  &4.24  &296.93  &0.77  &0.66 \\
    {{+ \sname (ours)}} & 675M & {2.4M} & [0, 0.85] &2.2   &1.40  &4.25  &293.57  &0.77  &0.67 \\
     \bottomrule
\end{tabular}
}
\end{table}

\begin{table}[h!]
\centering\small
\vspace{-9pt}
\caption{
    Performance comparison on ImageNet 512$\times$512 with CFG. 
    %REG uses CFG with $w=2.6$. 
}
\begin{tabular}{lcccccc}
    \toprule
    {\pz\pz Model} & Epochs  &  {\pz FID$\downarrow$} & {sFID$\downarrow$} & {IS$\uparrow$} & {Pre.$\uparrow$} & Rec.$\uparrow$ \\
    \arrayrulecolor{black}\midrule
    \multicolumn{7}{l}{\emph{Pixel diffusion}\vspace{0.02in}} \\
    \pz\pz {VDM$++$}~\cite{vdm}                   & -   & 2.65 & -    & 278.1 & -    & - \\
    \pz\pz {ADM-G, ADM-U}~\cite{ADM}      & 400 & 2.85 & 5.86 & 221.7 & 0.84 & 0.53 \\
    \pz\pz Simple diffusion (U-Net)~\cite{simple}    & 800 & 4.28 & -    & 171.0 & - & - \\
    \pz\pz Simple diffusion (U-ViT, L)~\cite{simple} & 800 & 4.53 & -    & 205.3 & - & - \\
    \arrayrulecolor{black!40}\midrule
    \multicolumn{7}{l}{\emph{Latent diffusion, Transformer}\vspace{0.02in}} \\
    \pz\pz MaskDiT~\cite{maskdit}                     & 800 & 2.50 & 5.10 & 256.3 & 0.83 & 0.56 \\ 
    \arrayrulecolor{black!30}\cmidrule(lr){1-7}
    \pz\pz {DiT-XL/2}~\cite{dit}        & 600 & 3.04 & 5.02 & 240.8 & 0.84 & 0.54  \\
    \arrayrulecolor{black!30}\cmidrule(lr){1-7}
    \pz\pz {SiT-XL/2}~\cite{sit}          & 600 & 2.62 & 4.18 & 252.2 & 0.84 & {0.57} \\
    {\pz\pz{+ REPA}}         &\pz80 & {2.44} & 4.21 & 247.3 &  0.84 & 0.56   \\
    {\pz\pz{+ REPA}}         & 100  &  {2.32} & {4.16} & {255.7} & {0.84} & {0.56}  \\
    {\pz\pz{+ REPA}}         & 200  & {2.08} & 4.19 & {274.6} & 0.83 & {0.58} \\
    %\hline
    {\pz\pz{\textbf{+ REG (ours)}}}     &\pz80 & {\textbf{1.68}} &\textbf{3.87}  &\textbf{306.9}  &0.80   &\textbf{0.63}    \\
    %{\pz\pz{+ REG}}         & 100  &  {} & {} & {} & {} & {}  \\
    %{\pz\pz{+ REG}}         & 200  & \textbf{} &  & {} &  & \textbf{} \\
    \arrayrulecolor{black}\bottomrule
\end{tabular}
%\vspace{-18pt}
\label{tab:512}
\end{table}

\section{512$\times$512 ImageNet}
To further validate REG's effectiveness, we conduct experiments at 512$\times$512 resolution following REPA's protocol~\cite{repa}. The RGB images are processed through the VAE~\cite{ldm} to yield 64$\times$64$\times$3 latents, with DINOv2~\cite{dinov2} providing both dense features and class token from 448$\times$448 inputs. 
As demonstrated in Tab.~\ref{tab:512}, REG surpasses the performance of REPA trained for 200 epochs and SiT-XL/2 trained for 600 epochs in terms of FID at only 80 epochs, demonstrating its superior effectiveness.

\section{Experimental setup}

\textbf{Hyperparameter setup.} Tab.~\ref{tab:hyper} presents the hyperparameter configurations of SiT + REG across different model scales. Following REPA's experimental protocol~\cite{repa}, we employ AdamW~\cite{adam,adam2} optimization with a batch size of $1\times10^{-4}$ and adopt DINOv2-B~\cite{dinov2} as the optimal alignment model, maintaining 250 denoising steps for all inference processes.

\textbf{CKNNA score.} Centered Kernel Nearest-Neighbor Alignment (CKNNA)~\cite{cknna} is a refined metric for measuring representational alignment between neural networks. It extends the traditional Centered Kernel Alignment (CKA)~\cite{cka} by focusing on local nearest-neighbor kernel alignment, emphasizing the similarity of local topological structures in representation spaces. In the following explanation, we largely adhere to the notation established in the original paper~\cite{cknna}. Mathematically, given two kernel matrices $K$ and $L$ derived from model representations, CKNNA computes alignment through three key steps. First, it applies a truncation function $\alpha(i, j)=1\left[\phi_j \in \operatorname{knn}\left(\phi_i\right) \wedge \psi_j \in \operatorname{knn}\left(\psi_i\right)\right]$, which selectively preserves pairwise relationships where sample $j$ is among the $k$-nearest neighbors of $i$ in both models. This creates a sparse kernel matrix that emphasizes local structures. Second, it calculates a weighted covariance of the truncated kernels:

\begin{equation}
\operatorname{Align}_{\mathrm{knn}}(K, L)=\sum_{i, j} \alpha(i, j) \cdot\left(\bar{K}_{i j} \cdot \bar{L}_{i j}\right),
\end{equation}

where $\bar{K}_{i j}$ and $\bar{L}_{i j}$ denote centered kernel values. Finally, the metric normalizes this alignment score to eliminate scale dependencies:

\begin{equation}
\operatorname{CKNNA}(K, L)=\frac{\operatorname{Align}_{\mathrm{knn}}(K, L)}{\sqrt{\operatorname{Align}_{\mathrm{knn}}(K, K) \cdot \operatorname{Align}_{\mathrm{knn}}(L, L)}} .
\end{equation}

We adopt REPA's CKNNA computation methodology~\cite{repa}, calculating scores exclusively between spatially averaged dense features from both the generative model and DINOv2-g representations~\cite{dinov2}. To ensure fair comparison, the class token is explicitly excluded from all CKNNA calculations.

\begin{table}[h!]
    \centering\small
    \vspace{-8pt}
    \caption{Hyperparameter settings across different model scales.}
    \resizebox{0.9\textwidth}{!}{% 
    \begin{tabular}{lccccc}
        \toprule
         \textbf{Backbone} & SiT-B & SiT-L & SiT-XL &  \\
        \midrule
        \textbf{Architecture} \\
        \#Params  &132M    &460M    &677M \\
        Input & 32$\times$32$\times$4 & 32$\times$32$\times$4 & 32$\times$32$\times$4  \\
        Layers  & 12 & 24 & 28  \\
        Hidden dim.  & 768 & 1,024 & 1,152  \\
        Num. heads  & 12 & 16 & 16  \\ 
        \midrule
        \textbf{REG settings} \\
        $\beta$  & 0.03 & 0.03 & 0.03  \\
        $\lambda$  & 0.5 & 0.5 & 0.5  \\
        Alignment depth  & 4 & 8 & 8  \\
        $\mathrm{sim}(\cdot, \cdot)$   & cos. sim. & cos. sim. & cos. sim.    \\
        Encoder $\mathcal{E}_{VF}(I)$  & DINOv2-B & DINOv2-B & DINOv2-B  \\
        \midrule
        \textbf{Optimization} \\
        Batch size  & 256 & 256 & 256 \\ 
        Optimizer  & AdamW & AdamW & AdamW \\
        lr  & 0.0001 &  0.0001 & 0.0001 \\
        $(\beta_1, \beta_2)$  & (0.9, 0.999) & (0.9, 0.999) & (0.9, 0.999) \\
        \midrule
        \textbf{Interpolants} \\
        $\alpha_t$  & $1-t$ & $1-t$ & $1-t$ \\
        $\sigma_t$  & $t$ & $t$ & $t$  \\
        $w_t$  & $\sigma_t$ & $\sigma_t$ & $\sigma_t$  \\
        Training objective  & v-prediction & v-prediction & v-prediction \\
        Sampler  & Euler-Maruyama & Euler-Maruyama & Euler-Maruyama  \\
        Sampling steps  & 250 & 250 & 250 \\
        \bottomrule
    \end{tabular}
    }
    \label{tab:hyper}
\end{table}

\section{More visualization results}
\begin{figure}[H]
    \centering
    \includegraphics[width=1.0\linewidth]{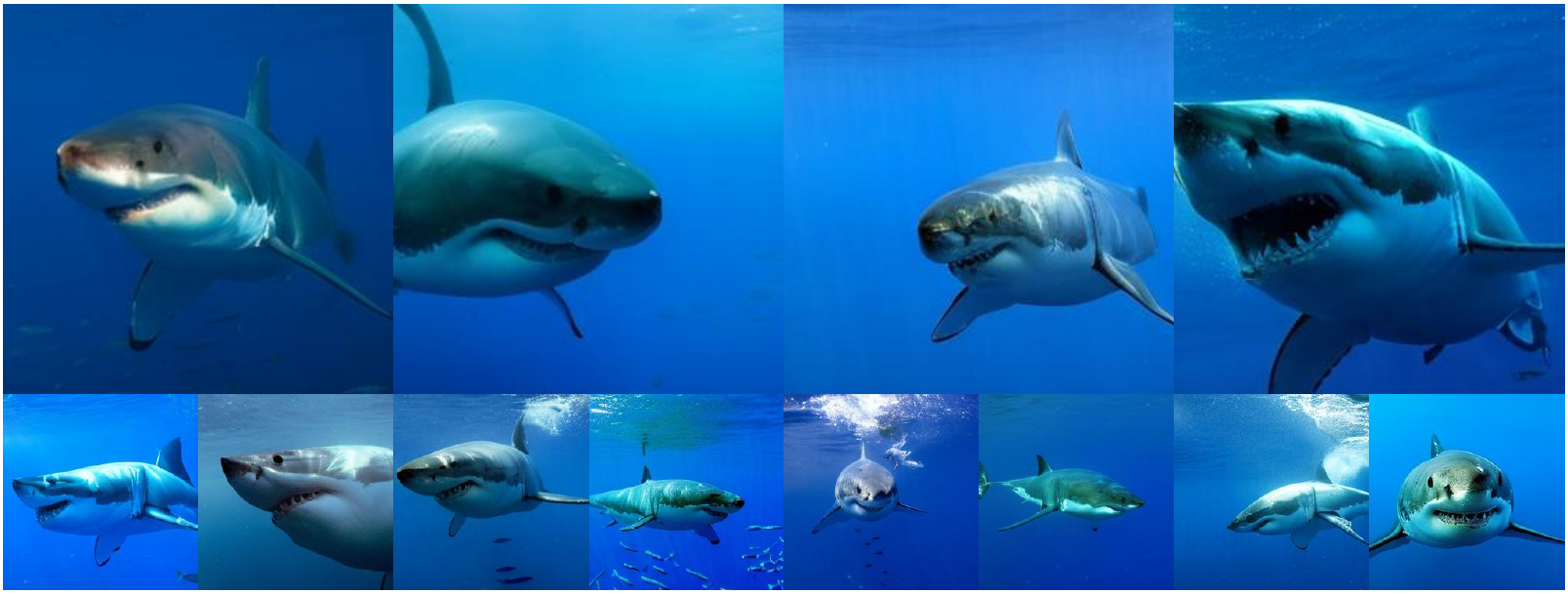}
    %\captionsetup{font={footnotesize}}
    \caption
    {
        The visualization results of SiT-XL/2 + REG use CFG with $w = 4.0$, and the class label is “Great white shark” (2).
    }
    \label{fig:app_com1}
    %\vspace{-20pt}
\end{figure}

\begin{figure}[H]
    \centering
    \includegraphics[width=1.0\linewidth]{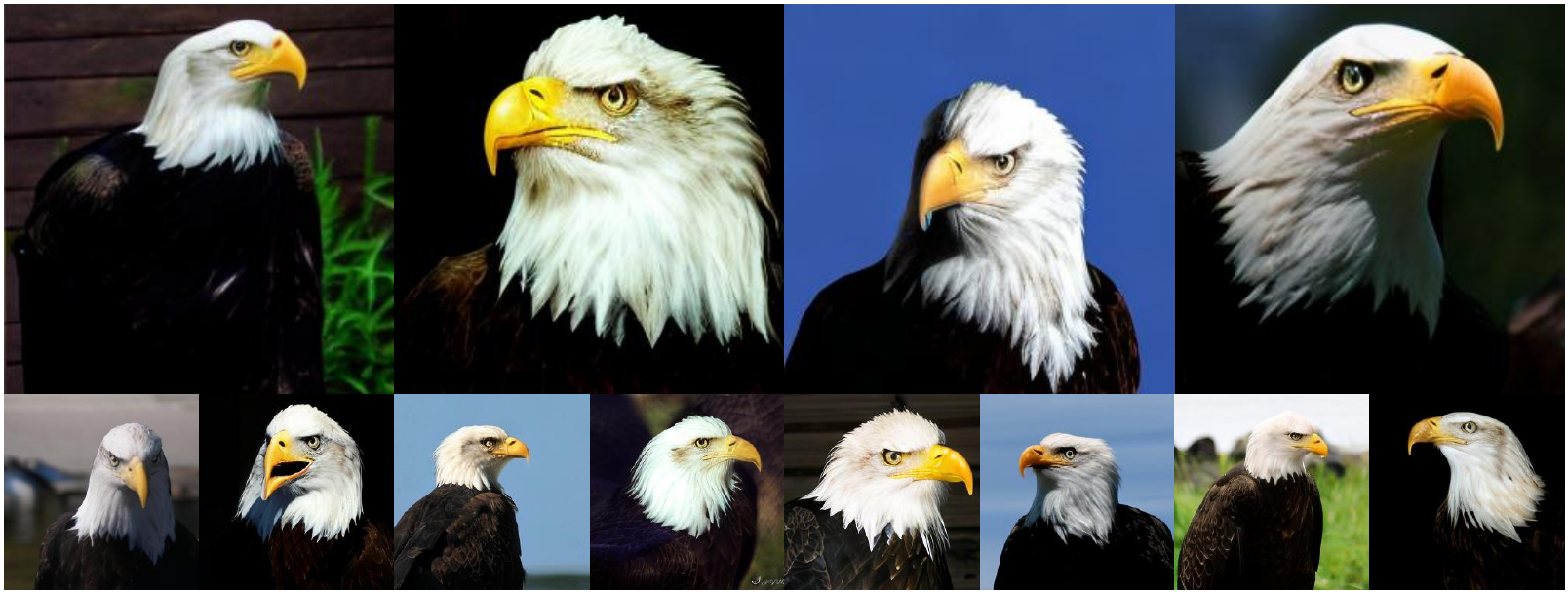}
    %\captionsetup{font={footnotesize}}
    \caption
    {
        The visualization results of SiT-XL/2 + REG use CFG with $w = 4.0$, and the class label is “Bald eagle” (22).
    }
    \label{fig:app_com2}
\end{figure}

\begin{figure}[H]
    \centering
    \includegraphics[width=1.0\linewidth]{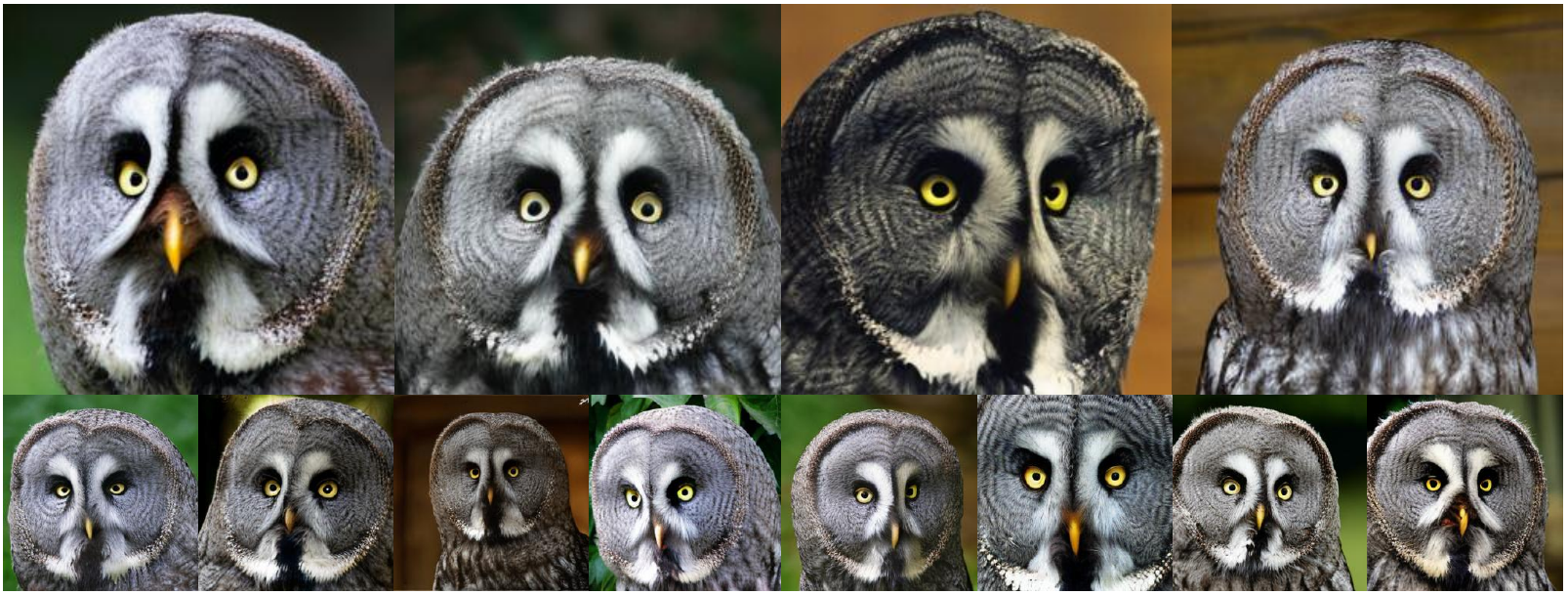}
    %\captionsetup{font={footnotesize}}
    \caption
    {
        The visualization results of SiT-XL/2 + REG use CFG with $w = 4.0$, and the class label is “Great grey owl” (24).
    }
    \label{fig:app_com3}
    %\vspace{-20pt}
\end{figure}

\begin{figure}[H]
    \centering
    \includegraphics[width=1.0\linewidth]{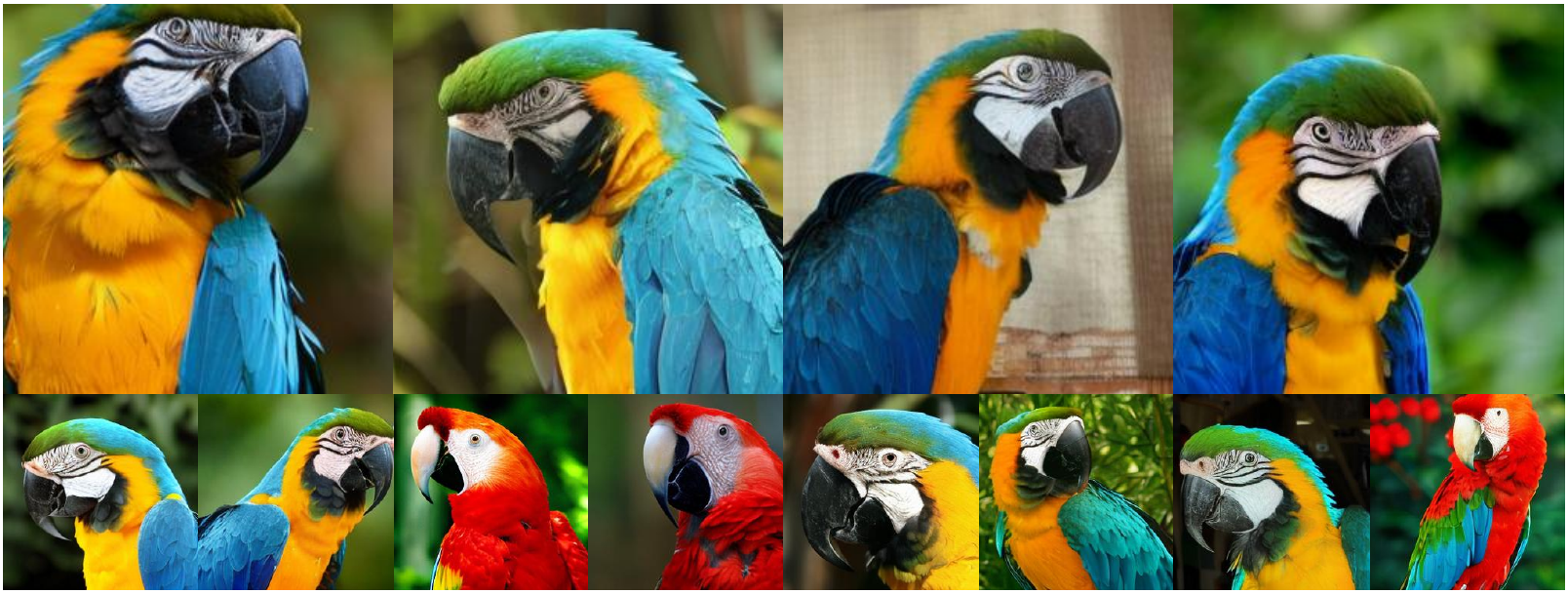}
    %\captionsetup{font={footnotesize}}
    \caption
    {
        The visualization results of SiT-XL/2 + REG use CFG with $w = 4.0$, and the class label is “Macaw” (88).
    }
    \label{fig:app_com4}
    %\vspace{-20pt}
\end{figure}

\begin{figure}[H]
    \centering
    \includegraphics[width=1.0\linewidth]{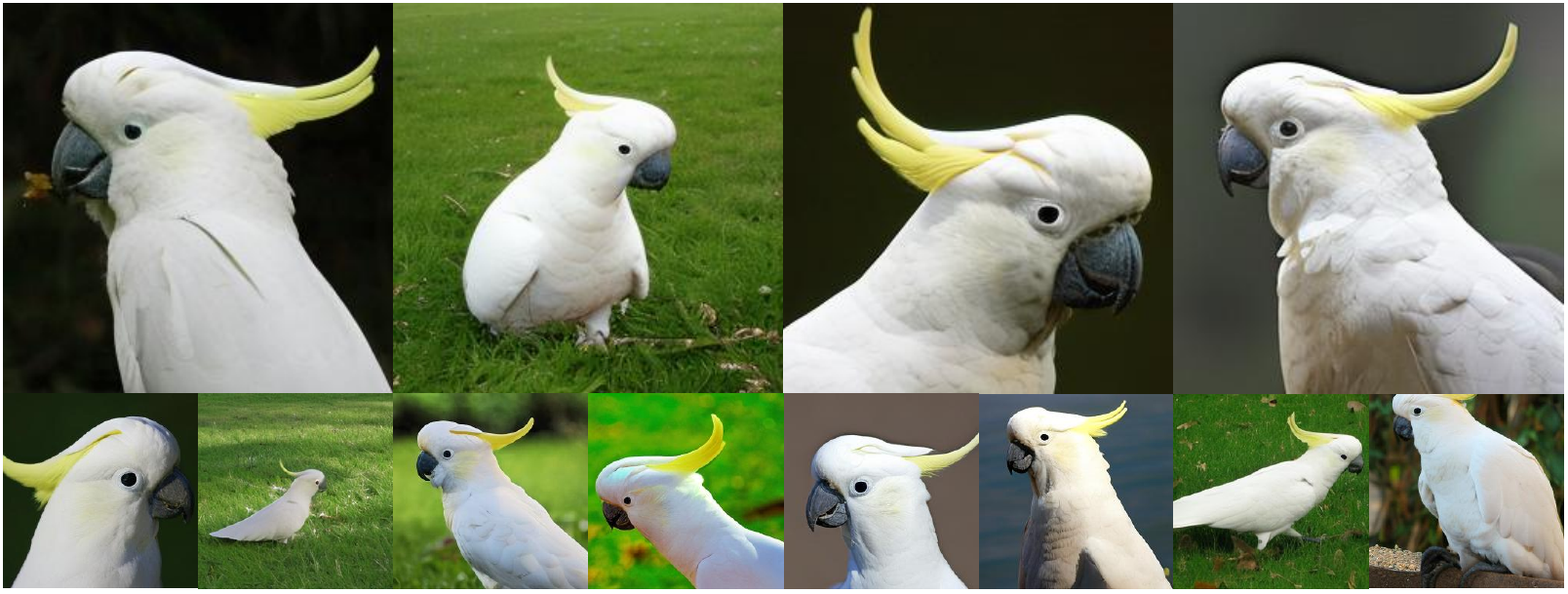}
    %\captionsetup{font={footnotesize}}
    \caption
    {
        The visualization results of SiT-XL/2 + REG use CFG with $w = 4.0$, and the class label is “Sulphur-crested cockatoo” (89).
    }
    \label{fig:app_com5}
    %\vspace{-20pt}
\end{figure}

\begin{figure}[H]
    \centering
    \includegraphics[width=1.0\linewidth]{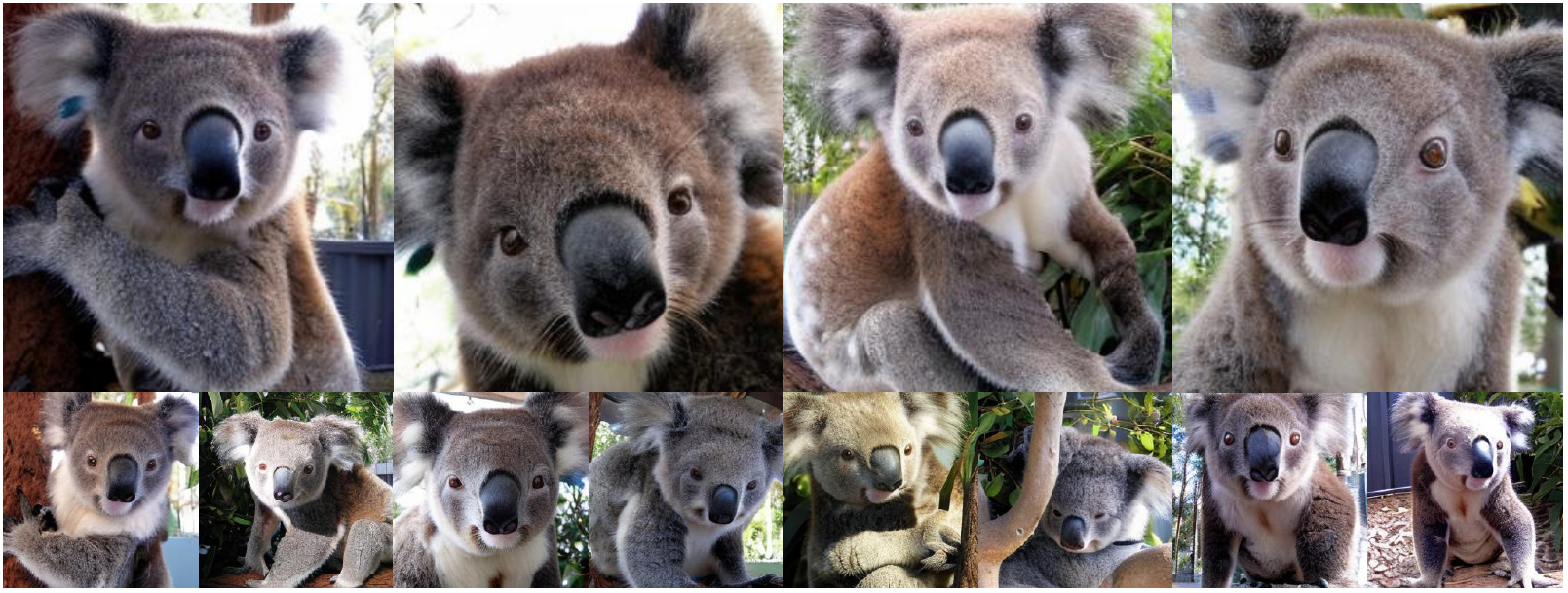}
    %\captionsetup{font={footnotesize}}
    \caption
    {
        The visualization results of SiT-XL/2 + REG use CFG with $w = 4.0$, and the class label is “Koala” (105).
    }
    \label{fig:app_com6}
    %\vspace{-20pt}
\end{figure}

\begin{figure}[H]
    \centering
    \includegraphics[width=1.0\linewidth]{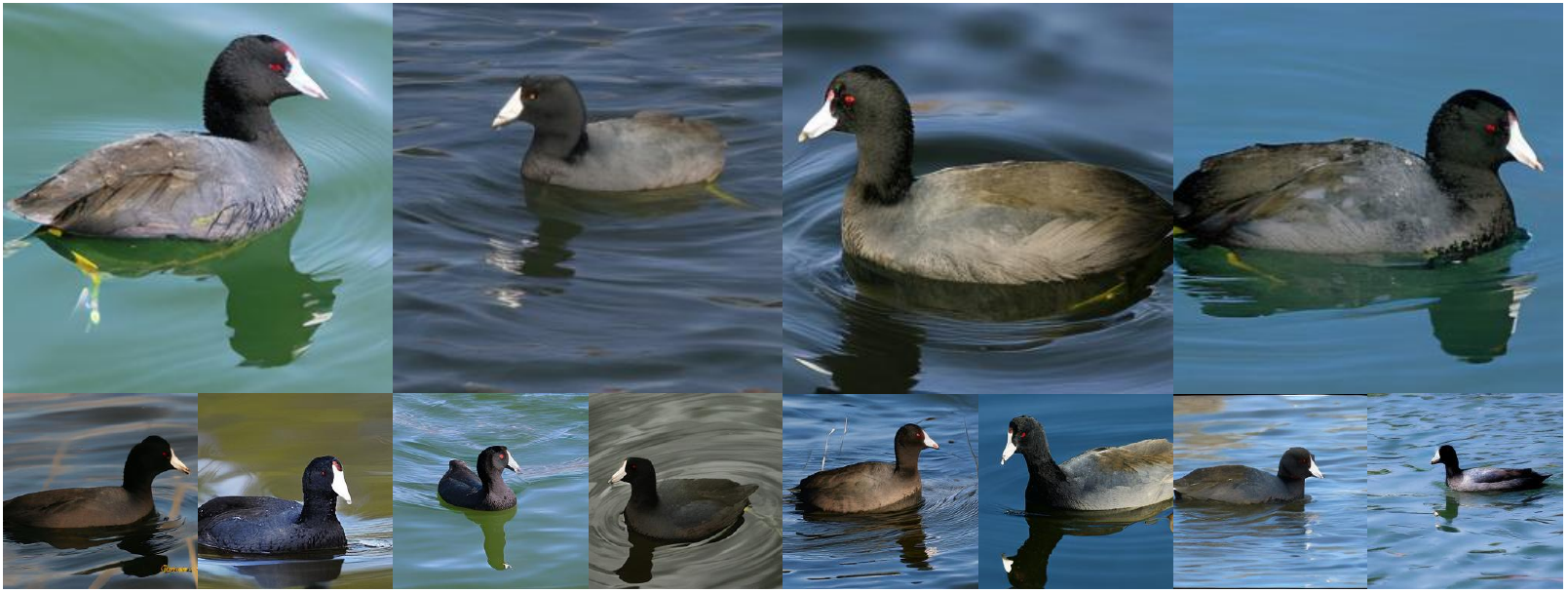}
    %\captionsetup{font={footnotesize}}
    \caption
    {
       The visualization results of SiT-XL/2 + REG use CFG with $w = 4.0$, and the class label is “American coot” (137).
    }
    \label{fig:app_com7}
\end{figure}

\begin{figure}[H]
    \centering
    \includegraphics[width=1.0\linewidth]{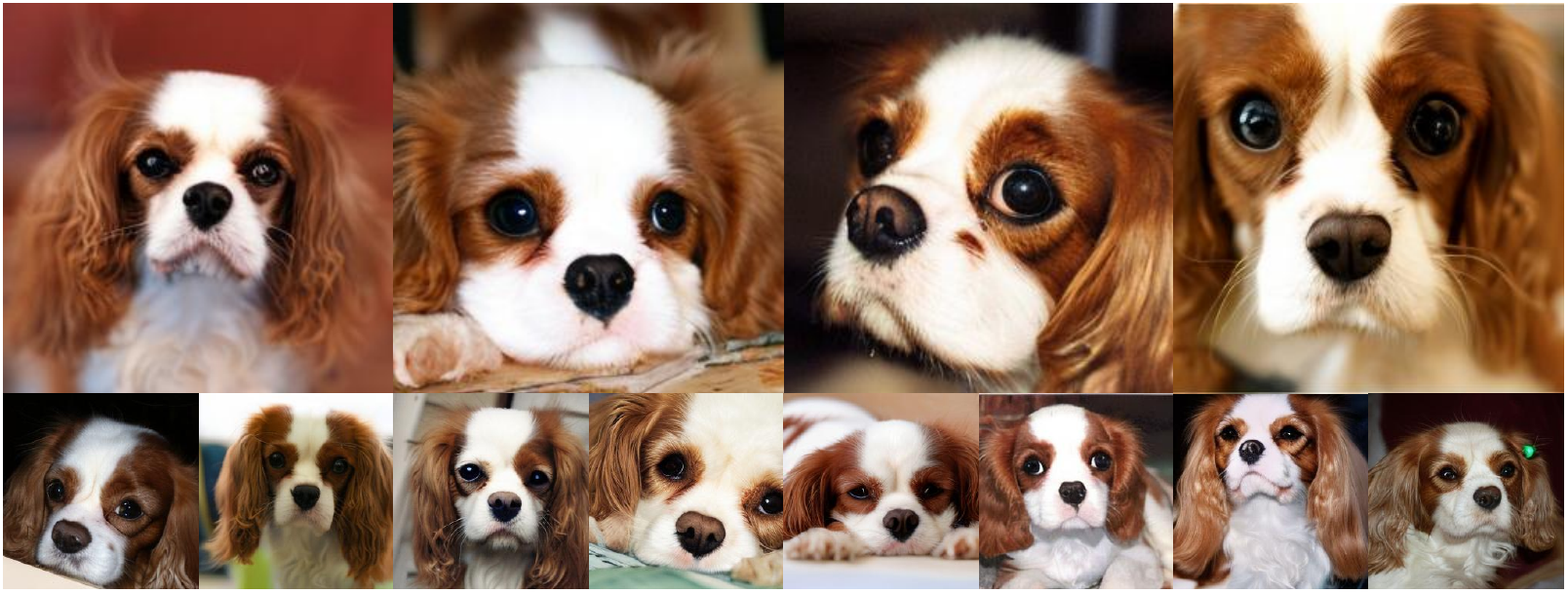}
    %\captionsetup{font={footnotesize}}
    \caption
    {
        The visualization results of SiT-XL/2 + REG use CFG with $w = 4.0$, and the class label is “Lesser panda” (156).
    }
    \label{fig:app_com8}
    %\vspace{-20pt}
\end{figure}

\begin{figure}[H]
    \centering
    \includegraphics[width=1.0\linewidth]{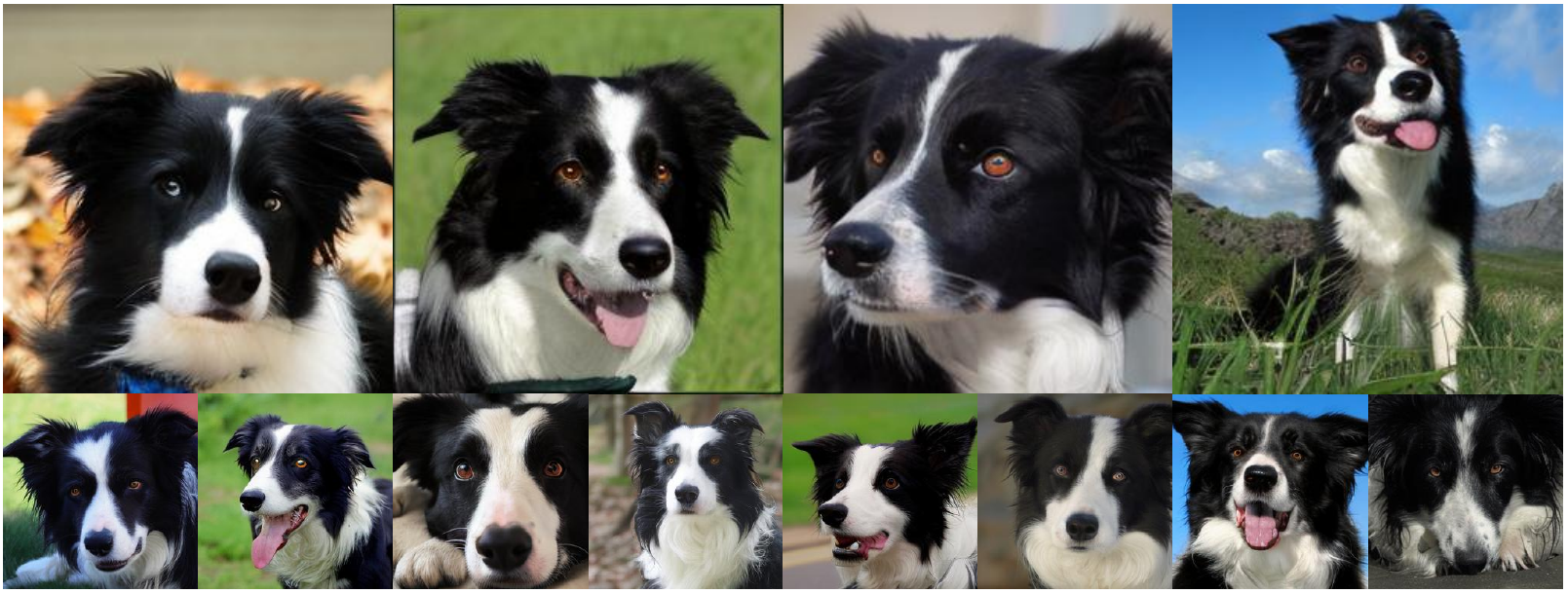}
    %\captionsetup{font={footnotesize}}
    \caption
    {
        The visualization results of SiT-XL/2 + REG use CFG with $w = 4.0$, and the class label is “Border collie” (232).
    }
    \label{fig:app_com9}
    %\vspace{-20pt}
\end{figure}

\begin{figure}[H]
    \centering
    \includegraphics[width=1.0\linewidth]{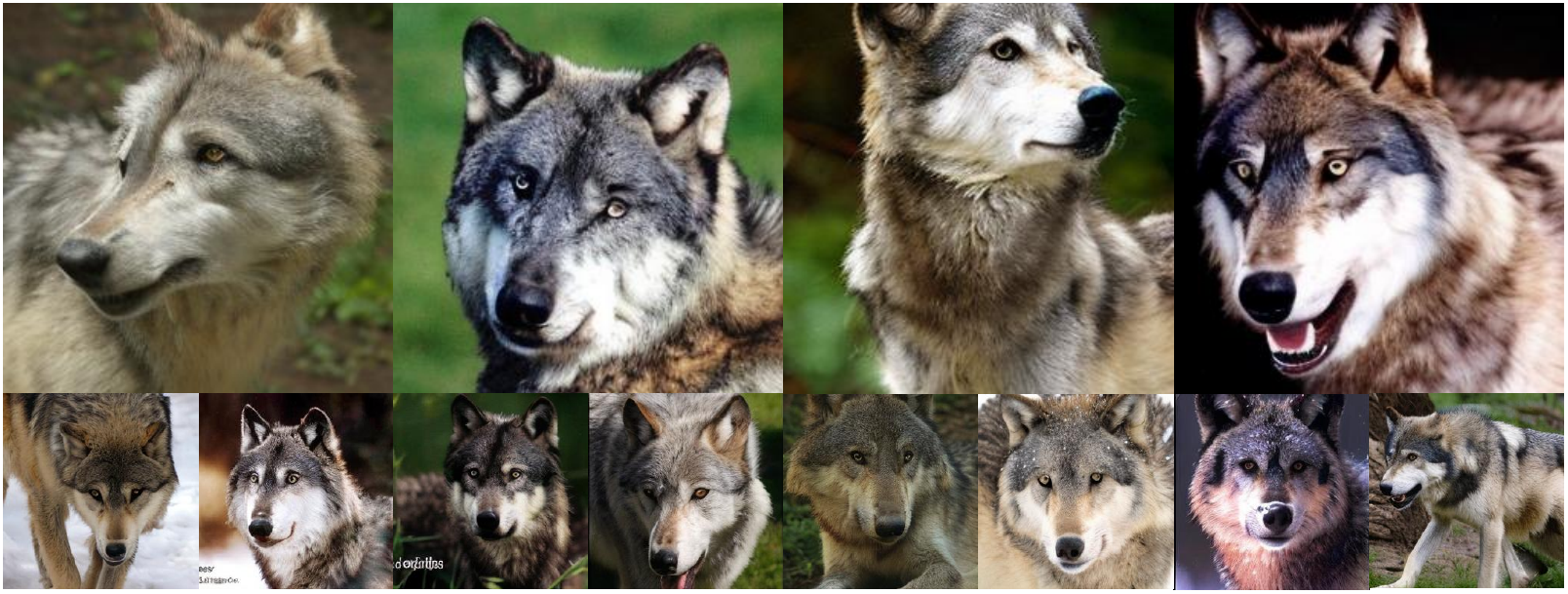}
    %\captionsetup{font={footnotesize}}
    \caption
    {
        The visualization results of SiT-XL/2 + REG use CFG with $w = 4.0$, and the class label is “Timber wolf” (269).
    }
    \label{fig:app_com10}
    %\vspace{-20pt}
\end{figure}

\begin{figure}[H]
    \centering
    \includegraphics[width=1.0\linewidth]{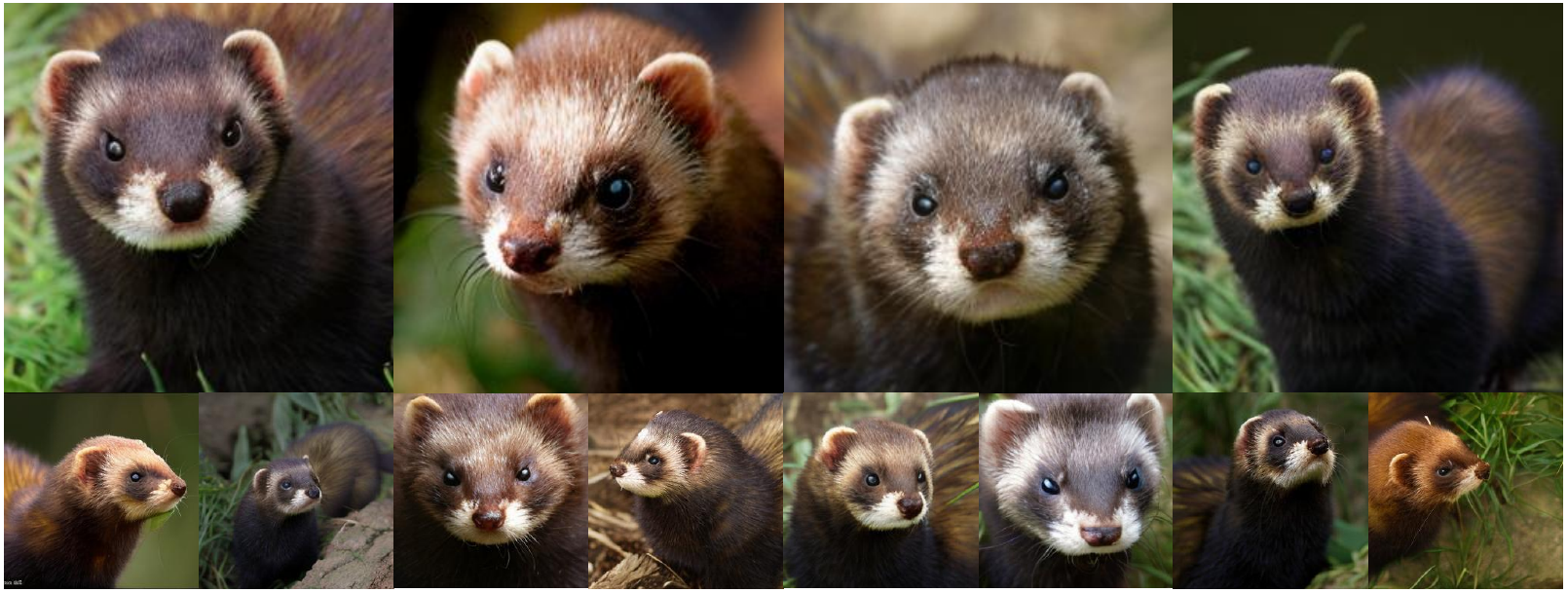}
    %\captionsetup{font={footnotesize}}
    \caption
    {
        The visualization results of SiT-XL/2 + REG use CFG with $w = 4.0$, and the class label is “Polecat” (358).
    }
    \label{fig:app_com11}
    %\vspace{-20pt}
\end{figure}

\begin{figure}[H]
    \centering
    \includegraphics[width=1.0\linewidth]{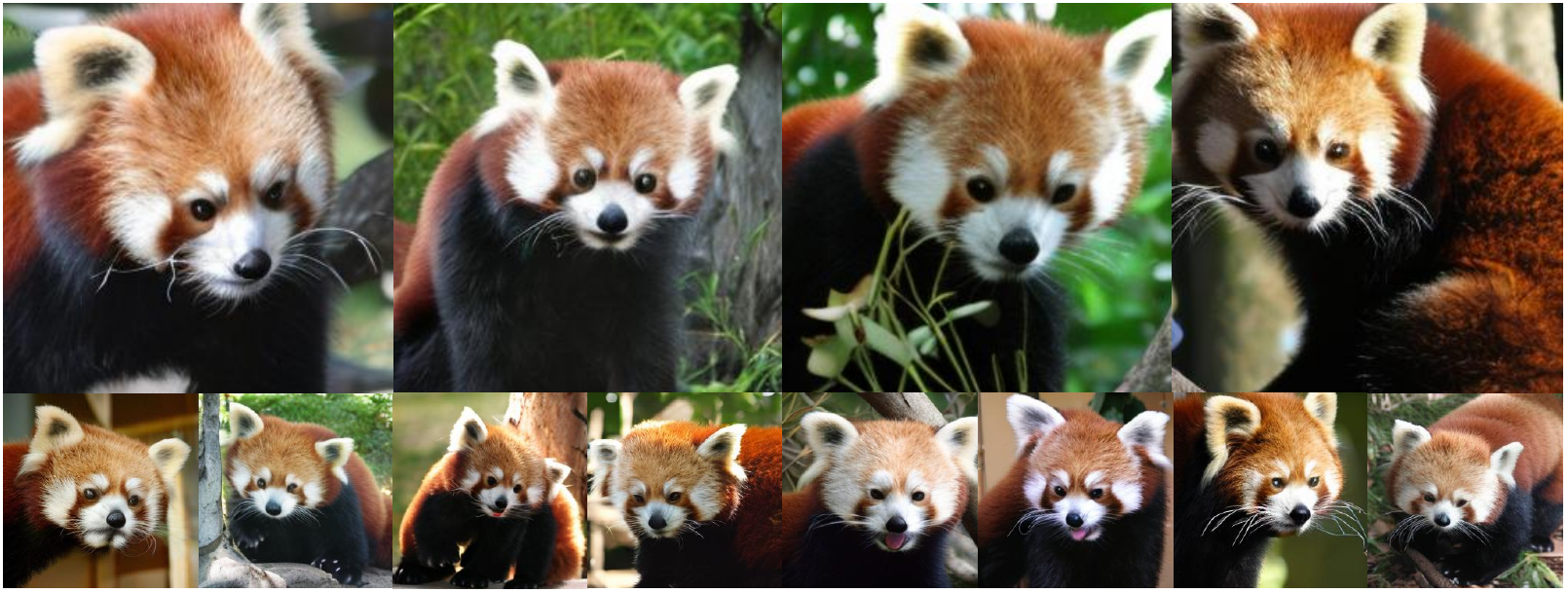}
    %\captionsetup{font={footnotesize}}
    \caption
    {
        The visualization results of SiT-XL/2 + REG use CFG with $w = 4.0$, and the class label is “Lesser panda” (387).
    }
    \label{fig:app_com12}
    %\vspace{-20pt}
\end{figure}

\begin{figure}[H]
    \centering
    \includegraphics[width=1.0\linewidth]{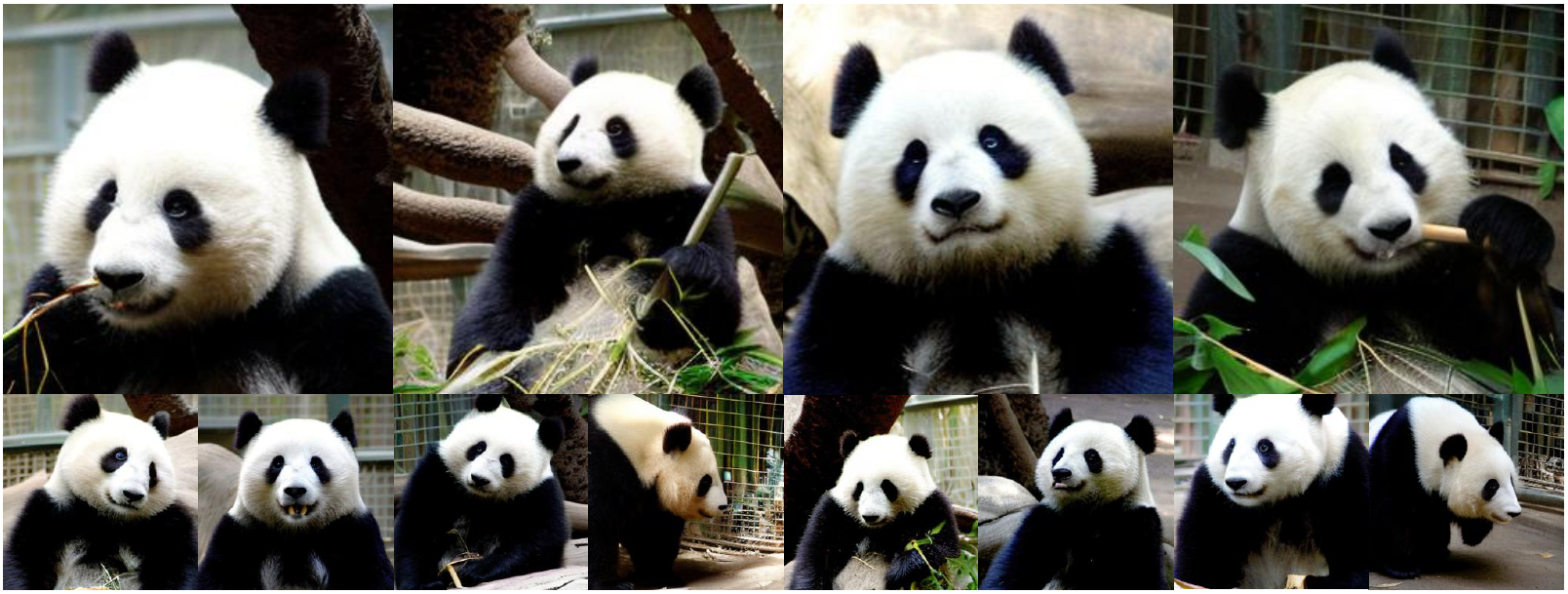}
    %\captionsetup{font={footnotesize}}
    \caption
    {
        The visualization results of SiT-XL/2 + REG use CFG with $w = 4.0$, and the class label is “Giant panda” (388).
    }
    \label{fig:app_com13}
    %\vspace{-20pt}
\end{figure}

\begin{figure}[H]
    \centering
    \includegraphics[width=1.0\linewidth]{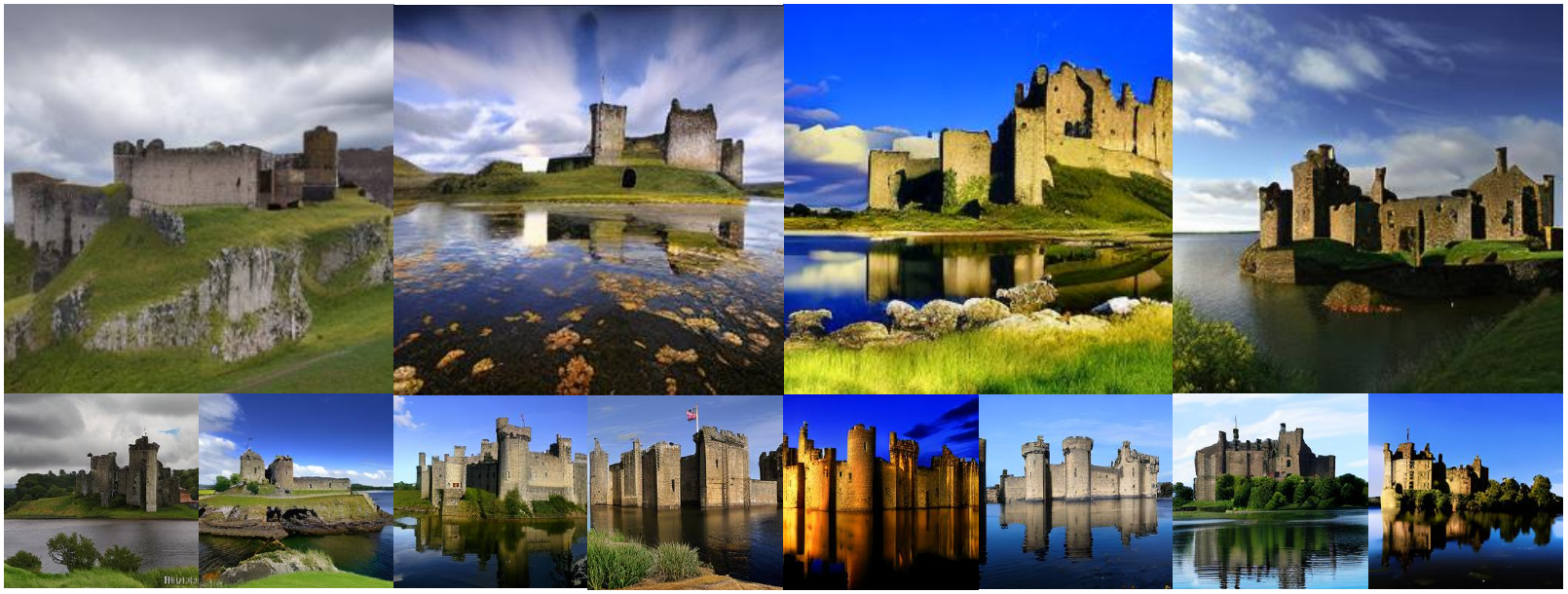}
    %\captionsetup{font={footnotesize}}
    \caption
    {
        The visualization results of SiT-XL/2 + REG use CFG with $w = 4.0$, and the class label is “Castle” (483).
    }
    \label{fig:app_com14}
    %\vspace{-20pt}
\end{figure}

\begin{figure}[H]
    \centering
    \includegraphics[width=1.0\linewidth]{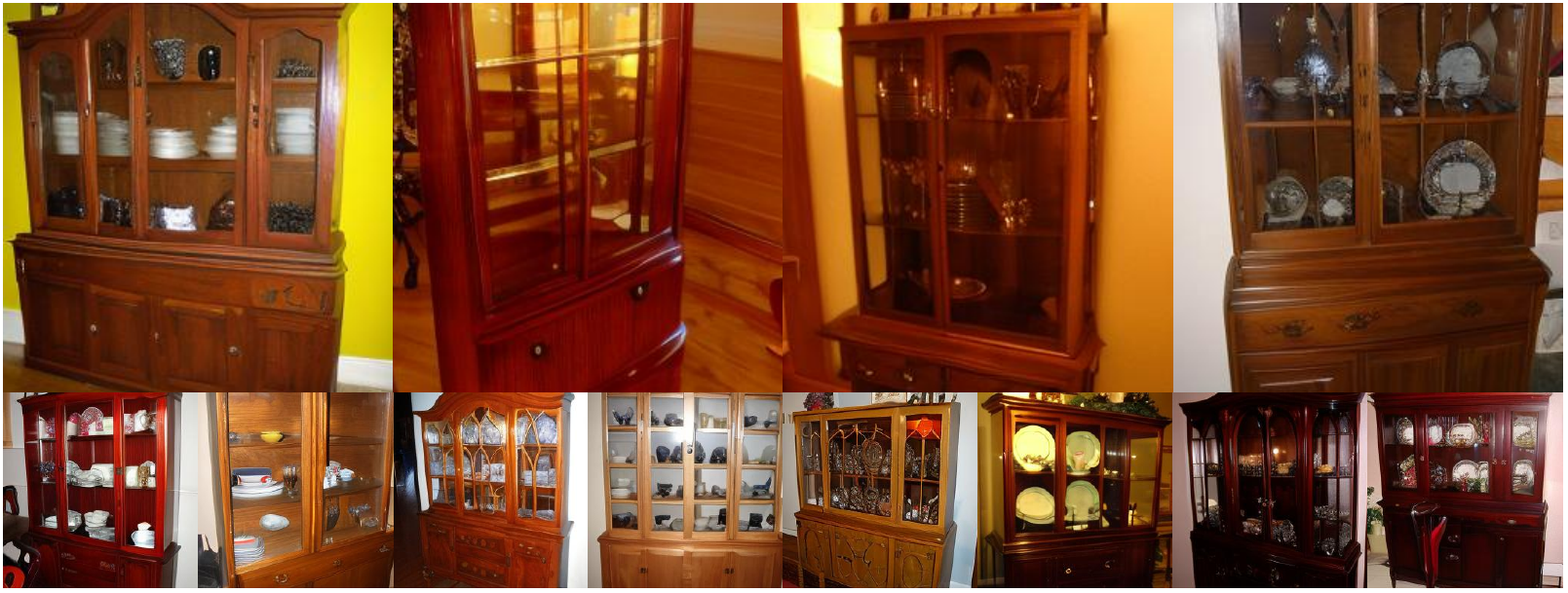}
    %\captionsetup{font={footnotesize}}
    \caption
    {
        The visualization results of SiT-XL/2 + REG use CFG with $w = 4.0$, and the class label is “China cabinet” (495).
    }
    \label{fig:app_com15}
    %\vspace{-20pt}
\end{figure}

\begin{figure}[H]
    \centering
    \includegraphics[width=1.0\linewidth]{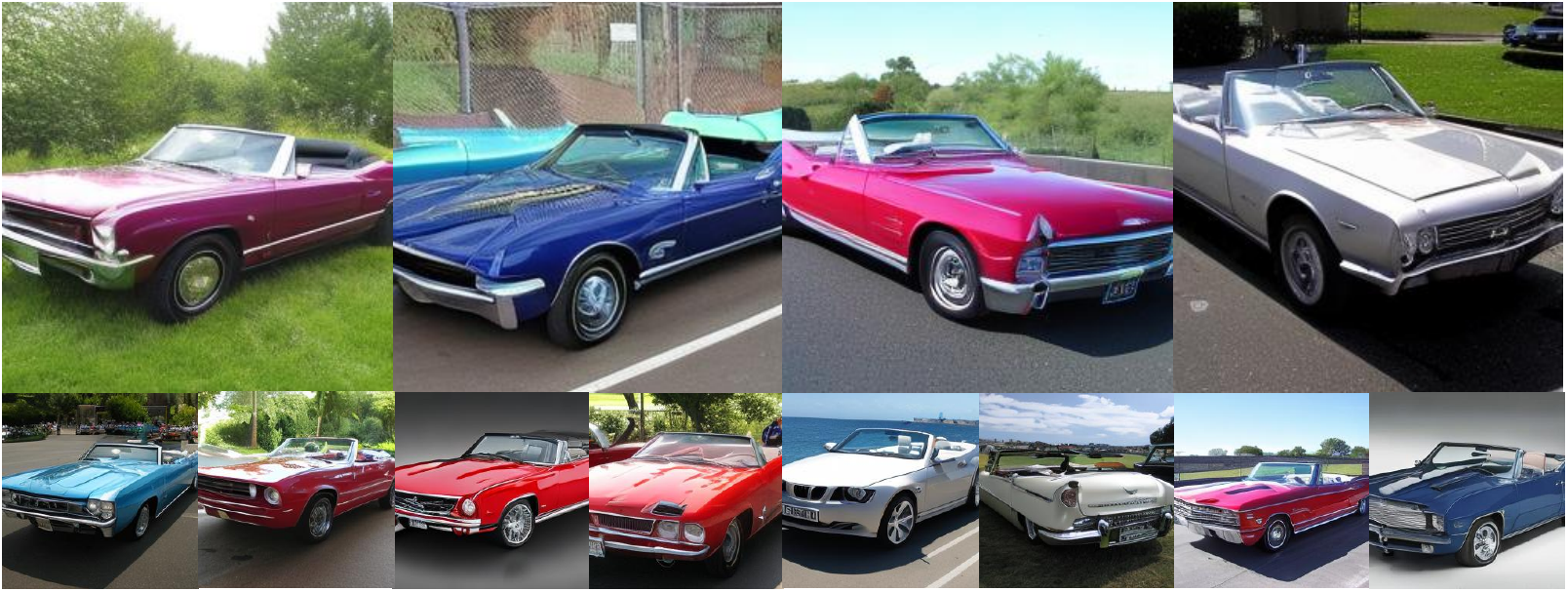}
    %\captionsetup{font={footnotesize}}
    \caption
    {
        The visualization results of SiT-XL/2 + REG use CFG with $w = 4.0$, and the class label is “Convertible” (511).
    }
    \label{fig:app_com16}
    %\vspace{-20pt}
\end{figure}

\begin{figure}[H]
    \centering
    \includegraphics[width=1.0\linewidth]{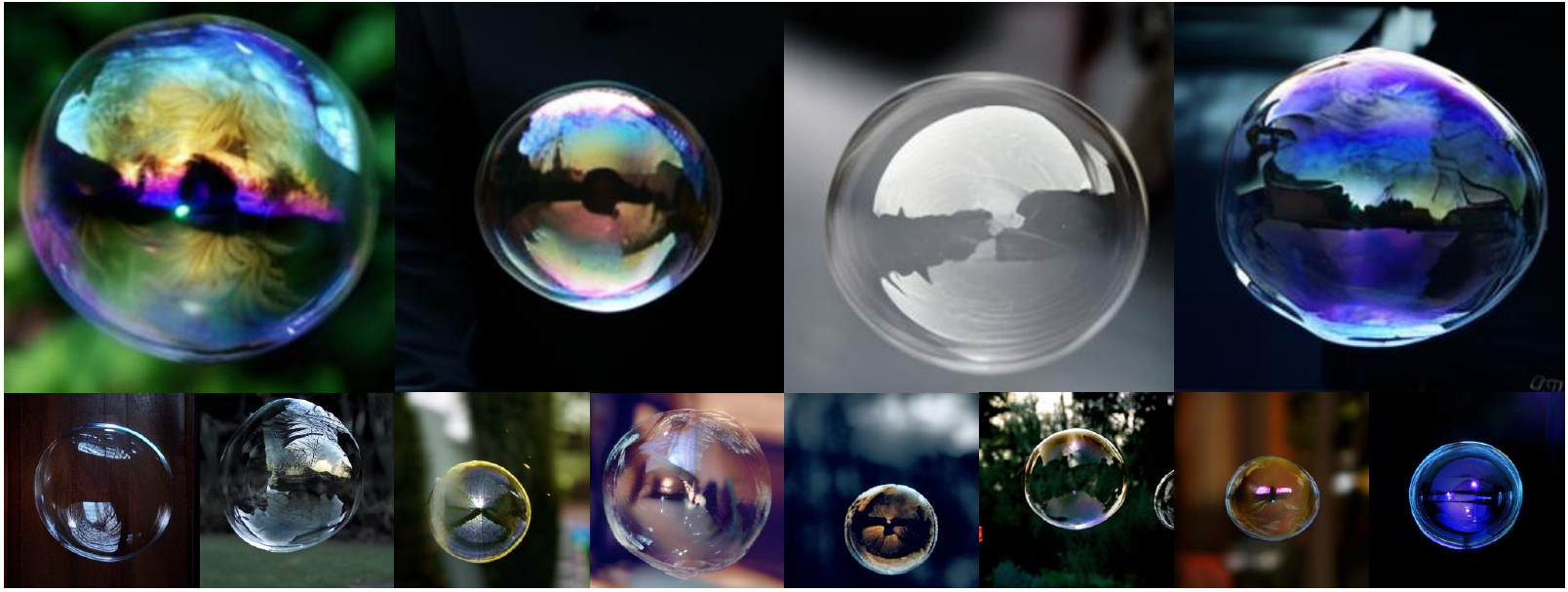}
    %\captionsetup{font={footnotesize}}
    \caption
    {
        The visualization results of SiT-XL/2 + REG use CFG with $w = 4.0$, and the class label is “Bubble” (971).
    }
    \label{fig:app_com17}
    %\vspace{-20pt}
\end{figure}

\begin{figure}[H]
    \centering
    \includegraphics[width=1.0\linewidth]{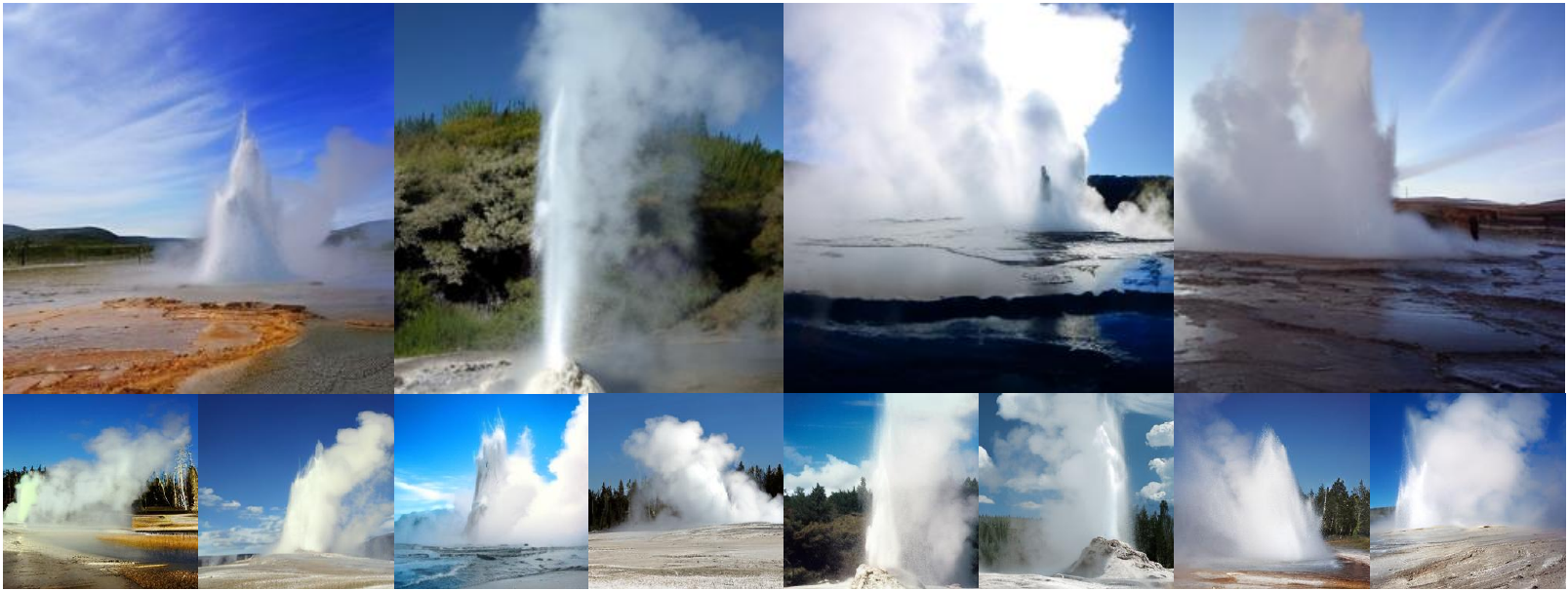}
    %\captionsetup{font={footnotesize}}
    \caption
    {
        The visualization results of SiT-XL/2 + REG use CFG with $w = 4.0$, and the class label is “Geyser” (974).
    }
    \label{fig:app_com18}
    %\vspace{-20pt}
\end{figure}

\begin{figure}[H]
    \centering
    \includegraphics[width=1.0\linewidth]{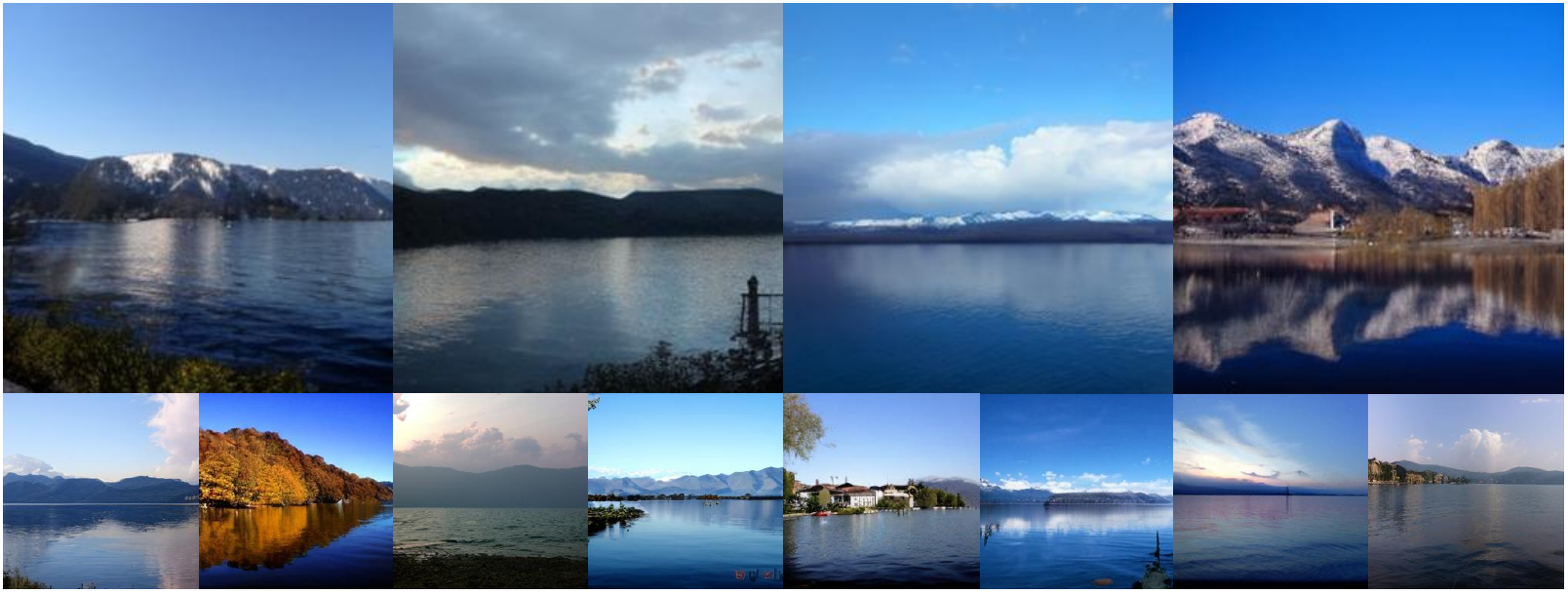}
    %\captionsetup{font={footnotesize}}
    \caption
    {
        The visualization results of SiT-XL/2 + REG use CFG with $w = 4.0$, and the class label is “Lakeside” (975).
    }
    \label{fig:app_com19}
    %\vspace{-20pt}
\end{figure}

\begin{figure}[H]
    \centering
    \includegraphics[width=1.0\linewidth]{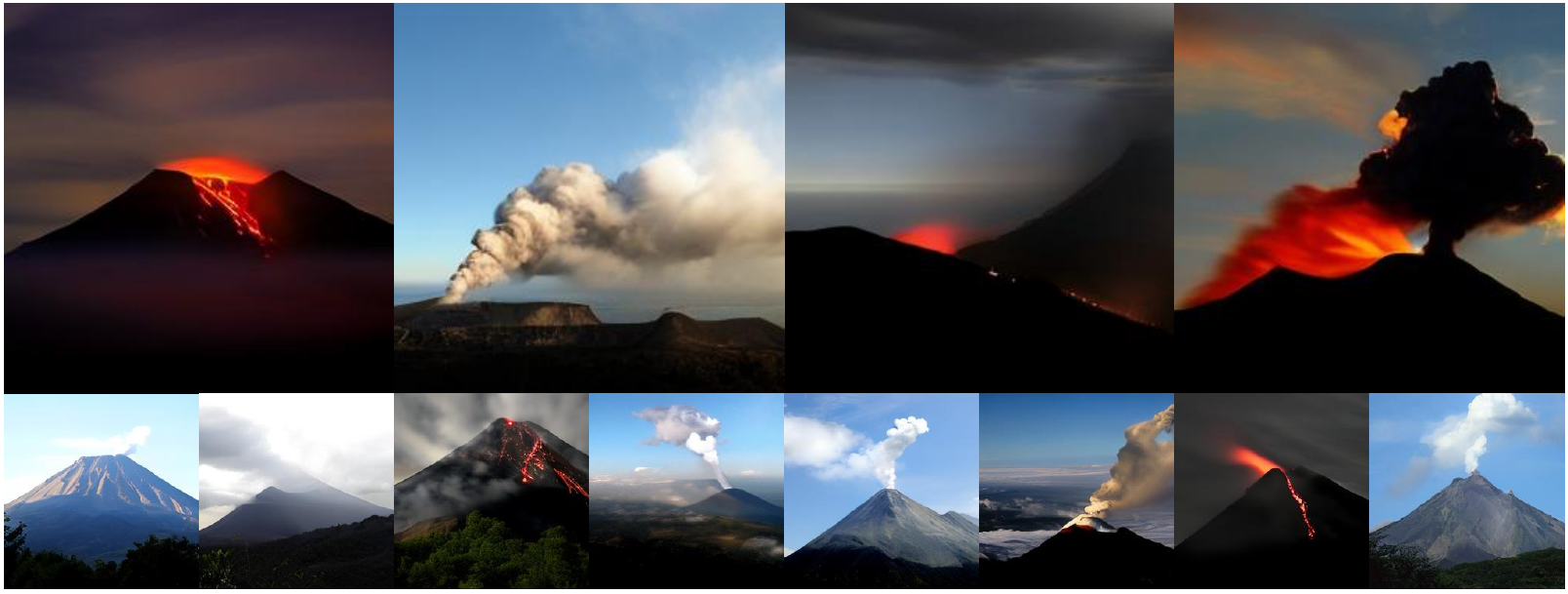}
    %\captionsetup{font={footnotesize}}
    \caption
    {
        The visualization results of SiT-XL/2 + REG use CFG with $w = 4.0$, and the class label is “Volcano” (980).
    }
    \label{fig:app_com20}
    %\vspace{-20pt}
\end{figure}

We present more visualization results of REG in Fig.~\ref{fig:app_com1} - \ref{fig:app_com20} with CFG ($w=4.0$).

\end{document}